\documentclass[journal]{IEEEtran}
\usepackage{cite}
\usepackage{amsmath,amssymb,amsfonts}
\usepackage{graphicx}

\graphicspath{ {./figs/} }
\usepackage{multirow}
\usepackage{subfigure}
\usepackage{algorithm}
\usepackage{algpseudocode}
\usepackage{amsmath}
\usepackage{amssymb}
\usepackage{enumitem}
\usepackage{booktabs}
\usepackage{multirow}
\usepackage{stfloats}
\usepackage{array}
\usepackage{xspace}
\usepackage{color}
\usepackage{epstopdf}
\usepackage{fancyhdr}  
\usepackage{array}

\usepackage{makecell}
\newcolumntype{C}[1]{>{\centering\arraybackslash}m{#1}}

\usepackage{booktabs}

\newcounter{itemlistc}

\def\BibTeX{{\rm B\kern-.05em{\sc i\kern-.025em b}\kern-.08em
    T\kern-.1667em\lower.7ex\hbox{E}\kern-.125emX}}

\IEEEoverridecommandlockouts  

\begin{document}

\title{Variational Matrix-Learning Fourier Networks for Parametric Multiphysics Surrogates}

\author{Xinyu~Li, 
        Jianhua~Zhang,
        and Liang~Chen,~\IEEEmembership{Member,~IEEE} 
\thanks{This work was supported in part by National Natural Science Foundation of China under Grant 92473105 and 62504151; in part by State Key Laboratory of Radio Frequency Heterogeneous Integration (Open Scientific Research Program No. KF2024005). \it{(Corresponding
        author: Liang Chen.)}}
\thanks{X.~Li, J.~Zhang, and L.~Chen are with the School of
Microelectronics, Shanghai University, Shanghai 201800, China (e-mail: lchenshu@shu.edu.cn).}
}



\maketitle

\begin{abstract}
Multiphysics simulation plays a critical role in system-technology co-optimization (STCO) for chiplet-based design. However, multiphysics phenomena are typically governed by partial differential equations (PDEs), whose repeated numerical solution is computationally expensive and memory intensive, especially in parametric design exploration. In this paper, we propose a variational matrix-learning Fourier network (VMLFN) framework for efficient parametric multiphysics surrogate modeling. First, a log-space sine neural-network representation is developed, in which frequency points are randomly sampled within a prescribed range in logarithmic space. A frequency-dependent decay factor is incorporated to regulate the contribution of different spectral components, and Dirichlet boundary conditions are embedded into the trial solution. The hidden-layer parameters are fixed, while only the output-layer weights are trainable. Second, the governing PDEs are reformulated in their variational weak forms, leading to a linear matrix system derived from the zero-gradient stationarity condition. The resulting formulation requires only first-order derivatives, avoids high-order automatic differentiation, and eliminates the need for penalty-coefficient tuning between PDE residuals and boundary conditions. The output weights can therefore be obtained efficiently through direct matrix solution. Finally, a heuristic scanning algorithm is introduced to determine a problem-adaptive maximum frequency, enabling the spectral basis to cover the dominant eigenfrequency range of the target problem. The proposed VMLFN is validated on three representative classes of multiphysics problems, including heat conduction, solid mechanics, and Helmholtz wave propagation. Experimental results across five benchmark cases demonstrate that the proposed framework achieves high accuracy and substantial computational speedup compared with conventional physics-informed neural networks and repeated finite-element simulations.
\end{abstract}
  

\section{Introduction}
\label{sec:introduction}
\IEEEPARstart{C}{hiplet}-based heterogeneous integration marks a pivotal transition in semiconductor design, shifting the industry from transistor-centric scaling toward system-level scaling~\cite{Su:ISSCC'23,Lau:TCPMT22}. This transition unlocks new opportunities for performance, modularity, and yield, but also elevates design complexity to an unprecedented level. In advanced chiplet systems, system-technology co-optimization (STCO) has emerged as a cornerstone methodology for next-generation electronic design automation, enabling coordinated optimization across circuits, architectures, packages, materials, and manufacturing technologies~\cite{Giacomini:JESSCDC24,Biswas:ISVLSI24}. At the heart of STCO lies multiphysics simulation, which serves as the physical foundation for predictive design and reliable optimization~\cite{Wang:IEDM24,Parekh:ICCD25}. Key analysis tasks in chiplet design are governed by fundamental partial differential equations. Electrostatic parasitic extraction is formulated via the Poisson equation~\cite{Abouelyazid:tcad22}, while thermal and mechanical integrity assessments require solving the heat-conduction and solid-mechanics equations~\cite{Zhu:DATE25}. Electromagnetic radiation and antenna effects are described by Helmholtz-type wave equations~\cite{Li:TGRS26}. These physics-based simulations are indispensable for capturing field interactions, identifying reliability risks, and guiding design decisions prior to costly fabrication. However, the computational burden of conventional multiphysics solvers poses a severe challenge to scalable STCO. Modern chiplet design demands repeated evaluations over large parametric spaces involving layout geometries, material stacks, power maps, and packaging configurations. Direct numerical simulation for each candidate design is often too expensive for rapid design-space exploration or optimization. Therefore, there is a pressing need for fast, accurate, and physics-consistent surrogate modeling techniques that can approximate parametric multiphysics responses while substantially reducing simulation cost.

Recently, artificial-intelligence (AI)-based surrogate modeling has rapidly advanced as a promising paradigm for accelerating multiphysics simulation, opening new opportunities for scientific computing and electronic design automation~\cite{Huang:TNNLS25,Long:ICML18,Lu:nmi21, li:ICLR21,Cao:NMI24,19PINN,Karniadakis:NRP'21,20DEM, Qi:TMTT25,Song:TGRS24,23binn,Wang:TNNLS25,Luca:TNNLS25,Zhang:TPAMI25,2020xpinn,20cpinn,20PIELM,Panghal:EC21,Sun:NPL19,21lelmm,2021elmforpde,23bayesianELM,24hdimELM,Chen:JML22,Davide:jcp26,Yang:SC20,Wang:IJCM26,Shivani:NS26,Chen:TCAD23,Wong:TAI24,Ren:25,Xiao:TII26,Ngom:SADM21,Chen:ICCAD23,Lu:SIAM21,Enrico:N21,Chen:TCAD25}. Existing AI surrogate models can be broadly classified into data-driven and physics-informed approaches. Purely data-driven models learn input-output mappings from large-scale simulation or experimental datasets and can provide fast inference once trained~\cite{Huang:TNNLS25,Long:ICML18,Lu:nmi21, li:ICLR21,Cao:NMI24}. However, their performance heavily depends on the quantity, quality, and coverage of the training data, making them vulnerable to poor generalization, limited extrapolation capability, and lack of physical consistency in unseen design regimes.

In contrast, physics-informed neural networks (PINNs) embed governing physical laws, typically expressed as partial differential equations (PDEs), together with boundary and initial conditions, directly into the learning objective~\cite{19PINN,Karniadakis:NRP'21,20DEM, Qi:TMTT25,Song:TGRS24,23binn,Wang:TNNLS25,Luca:TNNLS25,Zhang:TPAMI25,2020xpinn,20cpinn,20PIELM,Panghal:EC21,Sun:NPL19,21lelmm,2021elmforpde,23bayesianELM,24hdimELM,Chen:JML22,Davide:jcp26,Yang:SC20,Wang:IJCM26,Shivani:NS26,Chen:TCAD23,Wong:TAI24,Ren:25,Xiao:TII26,Ngom:SADM21,Chen:ICCAD23,Lu:SIAM21,Enrico:N21,Chen:TCAD25}. By enforcing physical residuals during training, PINNs provide a principled framework for reducing the dependence on labeled data while maintaining consistency with the underlying physics, making them attractive for complex multiphysics modeling tasks. Nevertheless, conventional PINNs still face several critical challenges in practical large-scale applications. Their training process is often computationally expensive, difficult to optimize, and prone to slow or unstable convergence, especially for coupled PDE systems, high-dimensional parametric spaces, and multiscale physical responses. In addition, the repeated evaluation of high-order derivatives through automatic differentiation imposes substantial memory and computational overhead. The limited representation efficiency of standard neural-network architectures further restricts their scalability and accuracy for large-scale multiphysics problems.

These limitations motivate the development of fast, data-efficient, and physics-consistent surrogate modeling techniques that can retain the physical rigor of PINNs while substantially improving training efficiency, convergence robustness, and scalability for parametric multiphysics simulation in STCO-oriented chiplet design. In this paper, we propose a novel Variational Matrix-Learning Fourier Networks for
Parametric Multiphysics Surrogates. Our contributions are outlined as follows:

\begin{itemize}

\item We propose a variational matrix-learning framework for solving PDE-governed multiphysics problems. Built upon the physics-informed learning paradigm, the proposed method introduces the variational form of the governing equations, thereby reducing the reliance on high-order automatic differentiation. By applying the zero-gradient stationarity condition, the optimization of trainable output weights is converted into an analytical matrix-solving problem. Moreover, the variational formulation alleviates the need to balance penalty coefficients between the governing-equation residuals and boundary-condition losses.

\item We develop a shallow log-space sine neural-network representation for parametric physical fields, inspired by the Fourier-series analytical solution structure. In the proposed architecture, the hidden-layer Fourier features are constructed using sine functions with randomly generated log-space frequencies. A frequency-dependent decay factor is introduced to regulate the contribution of different frequency components. Dirichlet boundary conditions are embedded directly into the Fourier neural network through fixed trial functions and envelope functions. The hidden-layer parameters remain fixed, while the output-layer coefficients are trainable and are efficiently determined through the proposed matrix-solving procedure.

\item We propose a heuristic scanning algorithm for determining a problem-adaptive maximum frequency, enabling the basis functions to cover the dominant eigenfrequency range of the target problem. The proposed method is applied to three representative classes of multiphysics problems, including heat conduction, solid mechanics, and Helmholtz wave propagation. Experimental results across five benchmark cases demonstrate that the proposed framework achieves high accuracy and substantial computational speedup compared with conventional PINNs and repeated finite-element-method simulations.
\end{itemize}

The remainder of this paper is organized as follows. Section~\ref{sec:RW} reviews related works. Section~\ref{sec:method} presents the proposed Variational Matrix-Learning Fourier Network framework and its formulation for heat conduction, solid mechanics, and Helmholtz problems. Section~\ref{sec:experimentalresults} reports the experimental results and analysis. Finally, Section~\ref{sec:concl} concludes this paper.

\section{Relevant Work}
\label{sec:RW}

Raissi {\it et al.} proposed physics-informed neural networks (PINNs)~\cite{19PINN}, an unsupervised learning framework for solving partial differential equations (PDEs) by incorporating the governing equations, boundary conditions, and initial conditions into the loss function. PINNs reduce the reliance on labeled data and provide a flexible paradigm for both forward and inverse PDE problems. Numerous studies have improved the performance of vanilla PINNs by designing advanced neural-network architectures, including the physics-informed Deep Operator Network (PIDON)~\cite{Qi:TMTT25}, physics-informed Fourier neural operator (PIFNO)~\cite{Song:TGRS24}, boundary-integral neural networks (BINNs)~\cite{23binn}, physics-informed dynamics representation learner (PIDO)~\cite{Wang:TNNLS25}, and physics-informed priors Bayesian neural networks (PIPs-BNNs)~\cite{Luca:TNNLS25}, Monte Carlo Neural PDE
Solver (MCNP Solver)~\cite{Zhang:TPAMI25}. In addition, domain decomposition strategies have been introduced into PINNs to enhance their approximation capability and scalability for large-scale or complex computational problems, such as extended PINNs (XPINNs)~\cite{2020xpinn} and conservative PINNs (cPINNs)~\cite{20cpinn}. Although PINNs alleviate the need for large labeled datasets, several challenges still hinder their broader applicability. Their training process is often computationally expensive, difficult to optimize, and prone to slow or unstable convergence, especially for coupled PDE systems, high-dimensional parametric spaces, and multiscale physical responses. Moreover, the repeated evaluation of high-order derivatives through automatic differentiation introduces substantial memory and computational overhead. The limited representation efficiency of standard neural-network architectures further restricts their scalability and accuracy in large-scale multiphysics applications.

\begin{figure*}[!hb]
    \centering
    \includegraphics[width=0.75\linewidth]{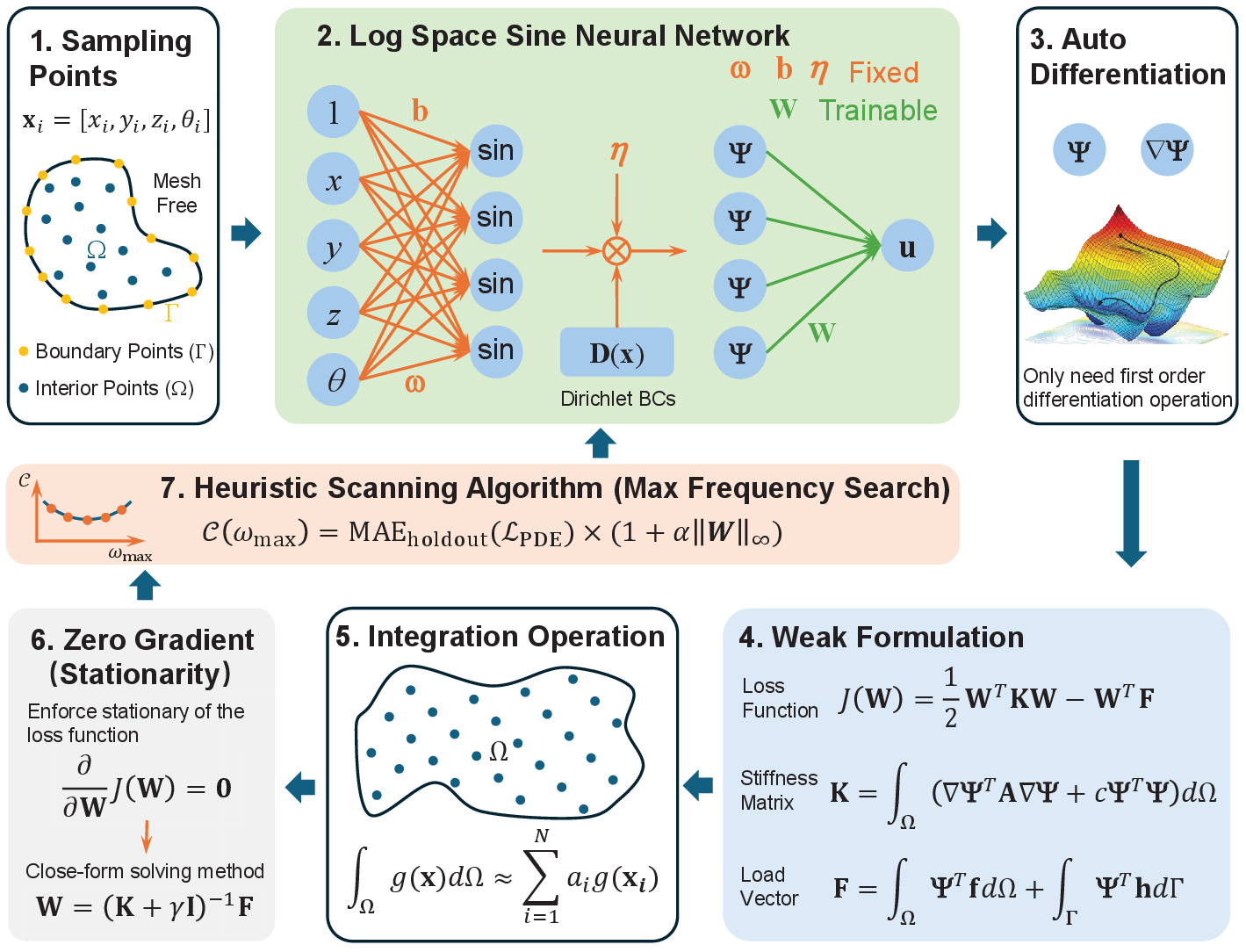}
    \caption{Overall framework of the proposed variational matrix-learning Fourier network for parametric multiphysics surrogate modeling. The framework consists of seven key steps: sampling-point generation, log-space sine neural network construction, automatic differentiation, weak-form formulation, numerical integration, zero-gradient matrix learning, and heuristic scanning algorithm.}
    \label{fig:VMLFN}
\end{figure*}

To improve the computational efficiency of PINNs, several studies have focused on reducing the costly training time caused by iterative back-propagation-based optimization. Among these efforts, back-propagation-free learning methods have emerged as an effective strategy, in which nonlinear optimization is replaced by direct matrix solvers. Dwivedi {\it et al.} introduced the extreme learning machine into the PINN framework and proposed the physics-informed extreme learning machine (PIELM) for solving PDEs~\cite{20PIELM}. Such methods can significantly accelerate the training procedure while retaining the mesh-free and physics-constrained characteristics of PINNs. Subsequently, several ELM-based and random-feature-based physics-informed methods have been developed for PDE applications~\cite{Panghal:EC21,Sun:NPL19,21lelmm,2021elmforpde,23bayesianELM,24hdimELM,Chen:JML22,Davide:jcp26,Yang:SC20,Wang:IJCM26,Shivani:NS26,Chen:TCAD23}. These studies have demonstrated the effectiveness of ELM-type solvers in reducing training costs and improving computational efficiency. However, because most existing methods employ shallow neural networks with randomly assigned hidden-layer parameters, their representation capability is relatively limited, and the resulting system matrices may suffer from severe ill-conditioning in terms of singular values. Consequently, achieving high accuracy remains challenging, particularly for complex PDEs with strong nonlinearity, high dimensionality, multiscale behavior, or complicated boundary conditions. In addition, PIELM-based methods usually rely on the strong form of PDEs and therefore require high-order differential operations, which may lead to unstable convergence and large approximation errors. Furthermore, the penalty coefficients associated with the governing equations, boundary conditions, and initial conditions are difficult to select and often require careful fine-tuning.


To improve the accuracy of PIELM-type methods, several studies have explored Fourier-related neural-network architectures. For example, sf-PINNs~\cite{Wong:TAI24,Ren:25} employ Fourier feature networks to accurately learn sinusoidal spaces. Physics-guided spectral neural networks (PGSNNs)\cite{Xiao:TII26} adopt a multiresolution spectral neural architecture to effectively represent intricate patterns and sharp gradients, leading to improved accuracy. Fourier series have also been used to represent PDE solutions\cite{Ngom:SADM21}. Chen {\it et al.} proposed ThermPINN for chip thermal simulation with sidewall boundary conditions, where discrete cosine neural networks (DCNs) are used to represent temperature distributions~\cite{Chen:ICCAD23}. Derived from the separation-of-variables method, DCNs can automatically enforce boundary conditions, such that the loss function only contains the governing equation. As a result, ThermPINN eliminates the need to determine optimal penalty coefficients for balancing the governing equation and boundary conditions. Hard-constraint techniques have also been introduced to consolidate multiple loss terms into a single loss function using the augmented Lagrangian method~\cite{Lu:SIAM21,Enrico:N21}. Based on DCNs, PISOV was further proposed to improve the training speed of ThermPINN by adopting an ELM strategy~\cite{Chen:TCAD25}. Nevertheless, the frequency selection of Fourier neural networks remains challenging. Chen {\it et al.} determine the frequency only by exploiting Neumann boundary conditions. In general, incorporating boundary conditions into Fourier neural networks is still nontrivial, which limits their flexibility and applicability to more general PDE problems.

In this paper, we propose a Variational Matrix-Learning Fourier Network for parametric multiphysics surrogate modeling. The proposed method derives the weak form of the governing PDE using a variational formulation and constructs a matrix-solving scheme to replace conventional ELM-type learning. Since only first-order derivatives are required, the proposed formulation improves numerical accuracy and stability while reducing the computational cost of derivative evaluation. Moreover, it avoids the need to balance penalty coefficients among the governing equations, boundary conditions, and initial conditions. Boundary conditions are incorporated through both the weak-form formulation and the proposed log-space sine neural network. Furthermore, we develop a log-space sine neural network, in which the frequencies are initialized using log-space parameters. A frequency-dependent decay factor is introduced to suppress oscillatory gradients and enhance training stability. Dirichlet boundary conditions are embedded directly into the neural-network representation. In addition, a heuristic scaling algorithm is proposed to determine the maximum frequency, thereby improving the adaptability and representation capability of the Fourier basis for complex parametric multiphysics problems.

\section{Methodology}
\label{sec:method}
In this section, we introduce the variational matrix-learning Fourier network (VMLFN) framework for solving second-order elliptic partial differential equations (PDEs), as illustrated in Fig.~\ref{fig:VMLFN}. First, a new Fourier neural-network architecture is developed, which incorporates log-space frequency sampling, a frequency-dependent decay factor, and embedded Dirichlet boundary conditions. Second, the variational principle is employed to derive the weak form of the governing PDEs. Based on the zero-gradient condition of the corresponding variational functional, a linear matrix system is obtained and solved using a matrix-learning strategy. Finally, a heuristic scanning algorithm is proposed to determine the maximum frequency used in the log-space sine neural network.

Subsequently, this unified framework is explicitly instantiated for three representative engineering physical fields: active Helmholtz wave propagation, steady-state heat conduction, and thermoelastic warping.

\subsection{Preliminary of second-order elliptic PDEs}
The multiphysics problems considered in this study can be mathematically formulated as a class of second-order elliptic partial differential equations (PDEs) defined over a bounded computational domain $\Omega$. Such a formulation provides a unified representation for a broad range of engineering physical fields, including steady-state heat conduction, acoustic Helmholtz propagation, and thermoelastic deformation. The generalized strong form is expressed as:
\begin{equation}
-\nabla \cdot (\mathbf{A}(\mathbf{x}) \nabla \mathbf{u}) + c(\mathbf{x}) \mathbf{u} = \mathbf{f}(\mathbf{x}), \quad \mathbf{x} \in \Omega
\end{equation}
where $\mathbf{u}$ denotes the unknown physical field, such as temperature, acoustic pressure, electric potential, or displacement. The coefficient tensor $\mathbf{A}(\mathbf{x})$ characterizes the material properties and may represent, for instance, the thermal conductivity tensor in heat conduction, the diffusion or stiffness tensor in elliptic mechanics problems, or other spatially varying anisotropic material parameters. The scalar coefficient $c(\mathbf{x})$ denotes the reaction, damping, or dissipation term, while $\mathbf{f}(\mathbf{x})$ represents the prescribed source, excitation, or body-force term. In addition, $\mathbf{n}$ is the outward unit normal vector on the boundary $\Gamma$.

The system is subject to mixed boundary conditions, including Dirichlet and Neumann boundary conditions, expressed as
\begin{equation}
\begin{aligned}
\mathbf{u} &= \mathbf{g}(\mathbf{x}), \quad \mathbf{x} \in \Gamma_D \quad \text{(Dirichlet Boundary)} \\
(\mathbf{A}(\mathbf{x}) \nabla \mathbf{u}) \cdot \mathbf{n} &= \mathbf{h}(\mathbf{x}), \quad \mathbf{x} \in \Gamma_N \quad \text{(Neumann Boundary)}
\end{aligned}
\end{equation}
where the Dirichlet boundary condition prescribes the value of the physical field on $\Gamma_D$, whereas the Neumann boundary condition specifies the flux, traction, or normal derivative response on $\Gamma_N$, depending on the physical interpretation of the governing equation.

\subsection{Log-Space Sine Neural Network}

\begin{figure}[!htb]
    \centering
    \includegraphics[width=\linewidth]{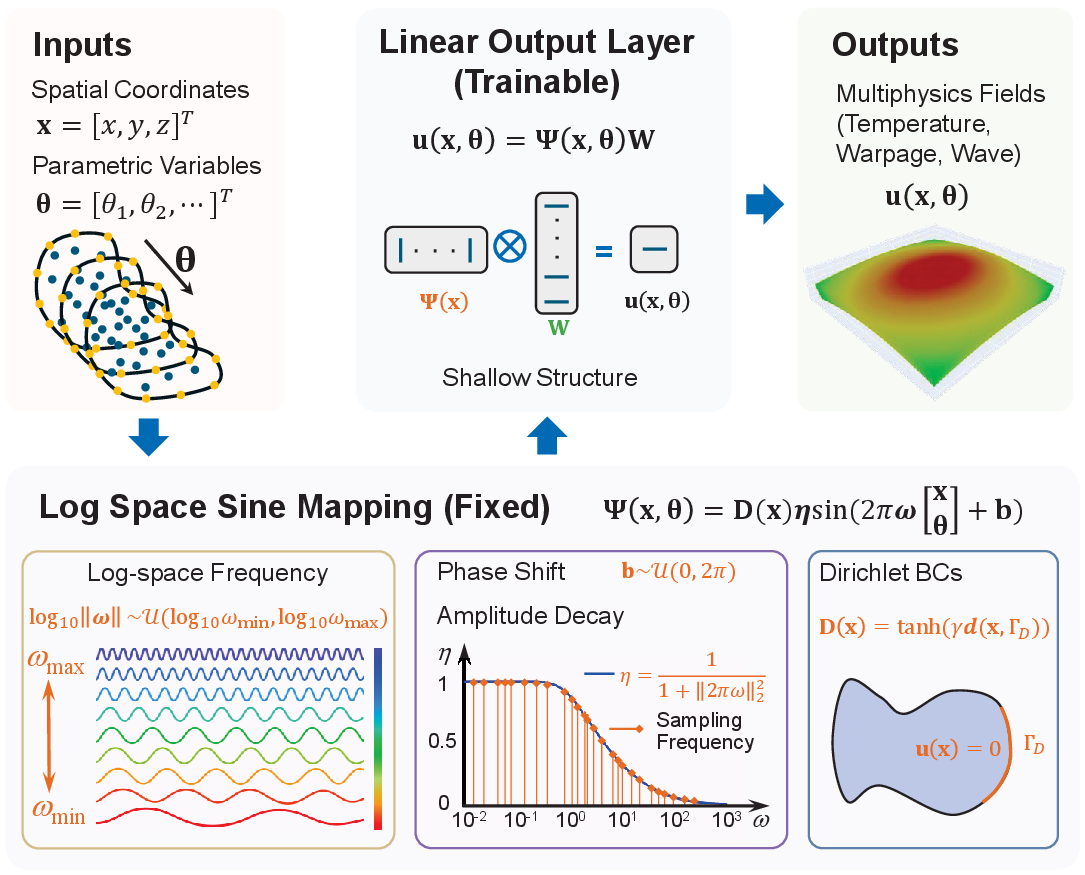}
    \caption{Log-space sine neural network.}
    \label{fig:FNN}
\end{figure}

To enhance the spectral representation capability of the proposed framework, we develop a new Fourier-type neural architecture, referred to as the log-space sine neural network, as shown in Fig.~\ref{fig:FNN}. Instead of using conventional trainable nonlinear layers, the proposed network represents the solution field as a linear combination of fixed sinusoidal basis functions. Specifically, the parametric solution of the governing PDE is approximated by
\begin{equation}
    \mathbf{u}(\mathbf{x},\boldsymbol{\theta})
    =
    \mathbf{\Psi}(\mathbf{x},\boldsymbol{\theta})\mathbf{W},
\end{equation}
where \(\mathbf{W}\) denotes the trainable coefficient matrix and \(\mathbf{\Psi}(\mathbf{x},\boldsymbol{\theta})\) is the constructed Fourier feature vector. The basis function is defined as
\begin{equation}
    \mathbf{\Psi}(\mathbf{x},\boldsymbol{\theta})
    =
    \mathbf{D}(\mathbf{x})\boldsymbol{\eta}
    \sin\left(
    2\pi\boldsymbol{\omega}
    \begin{bmatrix}
    \mathbf{x}\\
    \boldsymbol{\theta}
    \end{bmatrix}
    +\mathbf{b}
    \right),
\end{equation}
where \(\mathbf{x}\) denotes the spatial coordinates, including the \(x\)-, \(y\)-, and \(z\)-coordinates when applicable, and \(\boldsymbol{\theta}\) represents the parametric variables of the PDE, such as material properties, geometric parameters, excitation frequencies, or source locations. In the implementation, both the inputs \(\mathbf{x}\) and \(\boldsymbol{\theta}\), as well as the output field \(\mathbf{u}\), are normalized to \([-1,1]\) to improve numerical conditioning and ensure consistent scaling among different physical quantities.

The proposed log-space sine neural network consists of several key components, which are described as follows.

\textbf{Log-space frequency points \(\boldsymbol{\omega}\):}
The frequency matrix \(\boldsymbol{\omega}\) is fixed after initialization and is not updated during training. For each frequency vector, its direction is sampled isotropically from a unit hypersphere, while its magnitude is sampled from a logarithmic distribution. A typical sampling strategy is given by
\begin{equation}
    \log_{10}\left|\boldsymbol{\omega}\right|
    \sim
    \mathcal{U}
    \left(
    \log_{10}(\omega_{\min}),
    \log_{10}(\omega_{\max})
    \right),
\end{equation}
where \(\omega_{\min}\) and \(\omega_{\max}\) denote the minimum and maximum frequencies, respectively. In this work, \(\omega_{\min}\) can be set to a small positive value, e.g., \(0.05\), while \(\omega_{\max}\) is determined by the proposed heuristic scanning algorithm. Compared with uniformly sampled frequencies, the log-space strategy allocates basis functions over a broad spectral range in a more balanced manner. Low-frequency bases are used to capture global and smooth solution components, whereas high-frequency bases improve the representation of localized variations, sharp gradients, and multiscale responses. Therefore, this design alleviates the spectral-bias issue commonly observed in standard neural networks.

\textbf{Bias \(\mathbf{b}\):}
The bias vector \(\mathbf{b}\) is also fixed after initialization. Each component is independently sampled from a uniform distribution,
\begin{equation}
    \mathbf{b}\sim\mathcal{U}(0,2\pi).
\end{equation}
The random phase shift enriches the diversity of the sinusoidal basis functions and improves the translational flexibility of the representation. As a result, the network can approximate solution patterns with different spatial locations without requiring additional nonlinear transformations.

\textbf{Frequency-dependent decay factor \(\boldsymbol{\eta}\):}
Although high-frequency bases are important for representing multiscale or localized features, excessively large frequencies may generate highly oscillatory gradients and lead to ill-conditioned system matrices in the subsequent matrix-learning process. To mitigate this issue, a frequency-dependent decay factor is introduced as
\begin{equation}
    \boldsymbol{\eta} =
    \frac{1}{1+\left\|2\pi\boldsymbol{\omega}\right\|_2^2}.
\end{equation}
This factor attenuates the contribution of extremely high-frequency components and acts as a spectral regularizer. It suppresses pathological oscillations in the derivative terms while retaining sufficient approximation capability. Consequently, the resulting weak-form matrix system becomes better conditioned and more stable to solve.

\textbf{Dirichlet boundary constraint \(\mathbf{D}(\mathbf{x})\):}
To impose Dirichlet boundary conditions in a hard-constrained manner, a distance-based envelope function \(\mathbf{D}(\mathbf{x})\) is incorporated into the basis function. A typical choice is
\begin{equation}
    \mathbf{D}(\mathbf{x})
    =
    \tanh\left(
    \gamma d(\mathbf{x},\Gamma_D)
    \right),
\end{equation}
where \(d(\mathbf{x},\Gamma_D)\) denotes the distance from point \(\mathbf{x}\) to the Dirichlet boundary \(\Gamma_D\), and \(\gamma\) is a positive scaling parameter controlling the transition rate of the envelope. Since
\begin{equation}
    \mathbf{D}(\mathbf{x})=0,\quad \forall \mathbf{x}\in\Gamma_D,
\end{equation}
the basis function vanishes exactly on the Dirichlet boundary. Therefore, the homogeneous Dirichlet condition can be automatically satisfied without introducing additional penalty terms into the loss function.


Overall, the proposed log-space sine neural network combines random Fourier features, spectral scaling, and hard boundary enforcement into a unified representation. Since only the output coefficient matrix \(\mathbf{W}\) is trainable, the network is well suited for matrix-learning-based solvers. Moreover, by embedding boundary constraints directly into the basis and distributing frequencies in log space, the proposed architecture provides an efficient and stable representation for parametric multiphysics surrogate modeling.

\subsection{Variational Matrix-Learning Method}

To overcome the limitations of strong-form residual minimization, this work adopts a Ritz--Galerkin variational formulation and converts the original PDE problem into the minimization of an equivalent energy functional. Compared with strong-form PINN or PIELM formulations, the weak-form formulation only requires first-order spatial derivatives, naturally incorporates Neumann boundary conditions, and provides a principled route to construct a linear matrix-learning system.

Given an admissible test function \(\mathbf{v}\) satisfying \(\mathbf{v}=\mathbf{0}\) on \(\Gamma_D\), the strong-form equation is multiplied by \(\mathbf{v}\) and integrated over the computational domain \(\Omega\), yielding
\begin{equation}
\int_{\Omega}
\left[
-\nabla\cdot
\left(
\mathbf{A}\nabla\mathbf{u}
\right)
+
c\mathbf{u}
\right]^T
\mathbf{v}
\,d\Omega
=
\int_{\Omega}
\mathbf{f}^T\mathbf{v}
\,d\Omega .
\end{equation}
By applying integration by parts and the Gauss divergence theorem, the second-order differential operator is transferred to a first-order bilinear form as
\begin{equation}
\begin{split}
&\int_{\Omega}
\left[
(\nabla\mathbf{v})^T\mathbf{A}\nabla\mathbf{u}
+
c\mathbf{v}^T\mathbf{u}
\right]
d\Omega  \\
&\quad
-
\int_{\Gamma}
\mathbf{v}^T
\left(
\mathbf{A}\nabla\mathbf{u}
\right)\cdot\mathbf{n}
\,d\Gamma
=
\int_{\Omega}
\mathbf{f}^T\mathbf{v}
\,d\Omega .
\end{split}
\end{equation}
Since \(\mathbf{v}=\mathbf{0}\) on \(\Gamma_D\), and the Neumann boundary condition
\((\mathbf{A}\nabla\mathbf{u})\cdot\mathbf{n}=\mathbf{h}\) is prescribed on \(\Gamma_N\), the corresponding weak form is obtained as
\begin{equation}
\begin{split}
\int_{\Omega}
\left[
(\nabla\mathbf{v})^T\mathbf{A}\nabla\mathbf{u}
+
c\mathbf{v}^T\mathbf{u}
\right]
d\Omega
=
\int_{\Omega}
\mathbf{f}^T\mathbf{v}
\,d\Omega
+
\int_{\Gamma_N}
\mathbf{h}^T\mathbf{v}
\,d\Gamma .
\end{split}
\label{eq:weak_form}
\end{equation}

Equivalently, the solution can be obtained by finding the stationary point of the following energy functional:
\begin{equation}
\begin{split}
J(\mathbf{u})
=&
\frac{1}{2}
\int_{\Omega}
\left[
(\nabla\mathbf{u})^T\mathbf{A}\nabla\mathbf{u}
+
c\mathbf{u}^T\mathbf{u}
\right]
d\Omega  \\
&-
\int_{\Omega}
\mathbf{f}^T\mathbf{u}
\,d\Omega
-
\int_{\Gamma_N}
\mathbf{h}^T\mathbf{u}
\,d\Gamma .
\end{split}
\label{eq:general_functional}
\end{equation}
The stationary condition \(\delta J(\mathbf{u})=0\) is equivalent to the weak form in \eqref{eq:weak_form}. This variational formulation provides several advantages. First, it reduces the differentiability requirement of the trial solution from second-order differentiability to first-order differentiability, thereby lowering the computational cost and improving numerical stability. Second, the Neumann boundary condition is incorporated as a natural boundary condition through the boundary integral, avoiding additional penalty terms. Third, when \(\mathbf{A}\) is symmetric positive definite and \(c(\mathbf{x})\geq 0\), the functional is convex over the admissible space, which leads to a well-posed minimization problem.

In the proposed framework, the solution is represented by the log-space sine neural network. For notational simplicity, the trial solution is written as $\mathbf{u}(\mathbf{x},\boldsymbol{\theta})=
\boldsymbol{\Psi}(\mathbf{x},\boldsymbol{\theta})\mathbf{W}$,
where \(\boldsymbol{\Psi}\in\mathbb{R}^{1\times N_h}\) denotes the feature vector evaluated at \((\mathbf{x},\boldsymbol{\theta})\), \(N_h\) is the number of hidden Fourier basis functions, and \(\mathbf{W}\in\mathbb{R}^{N_h\times d_u}\) is the unknown coefficient matrix, with \(d_u\) being the dimension of the physical field. Substituting the trial solution into \eqref{eq:general_functional} transforms the original infinite-dimensional variational problem into a finite-dimensional quadratic minimization problem with respect to \(\mathbf{W}\):
\begin{equation}
\begin{split}
J(\mathbf{W})
=&
\frac{1}{2}
\int_{\Omega}
\left[
(\nabla\boldsymbol{\Psi}\mathbf{W})^T
\mathbf{A}
(\nabla\boldsymbol{\Psi}\mathbf{W})
+
c
(\boldsymbol{\Psi}\mathbf{W})^T
(\boldsymbol{\Psi}\mathbf{W})
\right]
d\Omega  \\
&-
\int_{\Omega}
\mathbf{f}^T
(\boldsymbol{\Psi}\mathbf{W})
\,d\Omega
-
\int_{\Gamma_N}
\mathbf{h}^T
(\boldsymbol{\Psi}\mathbf{W})
\,d\Gamma .
\end{split}
\end{equation}

By collecting the terms associated with \(\mathbf{W}\), the functional can be rewritten in the standard quadratic form
\begin{equation}
J(\mathbf{W})
=
\frac{1}{2}
\operatorname{tr}
\left(
\mathbf{W}^T\mathbf{K}\mathbf{W}
\right)
-
\operatorname{tr}
\left(
\mathbf{W}^T\mathbf{F}
\right),
\label{eq:quadratic_energy}
\end{equation}
where \(\mathbf{K}\in\mathbb{R}^{N_h\times N_h}\) is the global stiffness matrix and \(\mathbf{F}\in\mathbb{R}^{N_h\times d_u}\) is the global load matrix, defined as
\begin{equation}
\begin{aligned}
\mathbf{K}
&=
\int_{\Omega}
\left[
(\nabla\boldsymbol{\Psi})^T
\mathbf{A}
\nabla\boldsymbol{\Psi}
+
c\boldsymbol{\Psi}^T\boldsymbol{\Psi}
\right]
d\Omega, \\
\mathbf{F}
&=
\int_{\Omega}
\boldsymbol{\Psi}^T\mathbf{f}
\,d\Omega
+
\int_{\Gamma_N}
\boldsymbol{\Psi}^T\mathbf{h}
\,d\Gamma .
\end{aligned}
\label{eq:KF_def}
\end{equation}
Here, the trace operator is used to consistently describe scalar energy for vector-valued physical fields.

According to the Rayleigh--Ritz condition, the equilibrium solution is obtained by enforcing the stationary condition of \eqref{eq:quadratic_energy} with respect to the coefficient matrix \(\mathbf{W}\):
\begin{equation}
\frac{\partial J}{\partial \mathbf{W}}
=
\mathbf{K}\mathbf{W}
-
\mathbf{F}
=
\mathbf{0}.
\end{equation}
Therefore, the optimal coefficient matrix is obtained from the following linear system:
\begin{equation}
\mathbf{K}\mathbf{W}
=
\mathbf{F}.
\label{eq:linear_system}
\end{equation}
This formulation eliminates the need for iterative back-propagation-based optimization and avoids issues such as learning-rate tuning, vanishing gradients, and slow convergence. In addition, because the unknowns appear linearly in the trial solution, the resulting learning problem is reduced to solving a deterministic matrix system.


\begin{figure}[!htb]
    \centering
    \includegraphics[width=\linewidth]{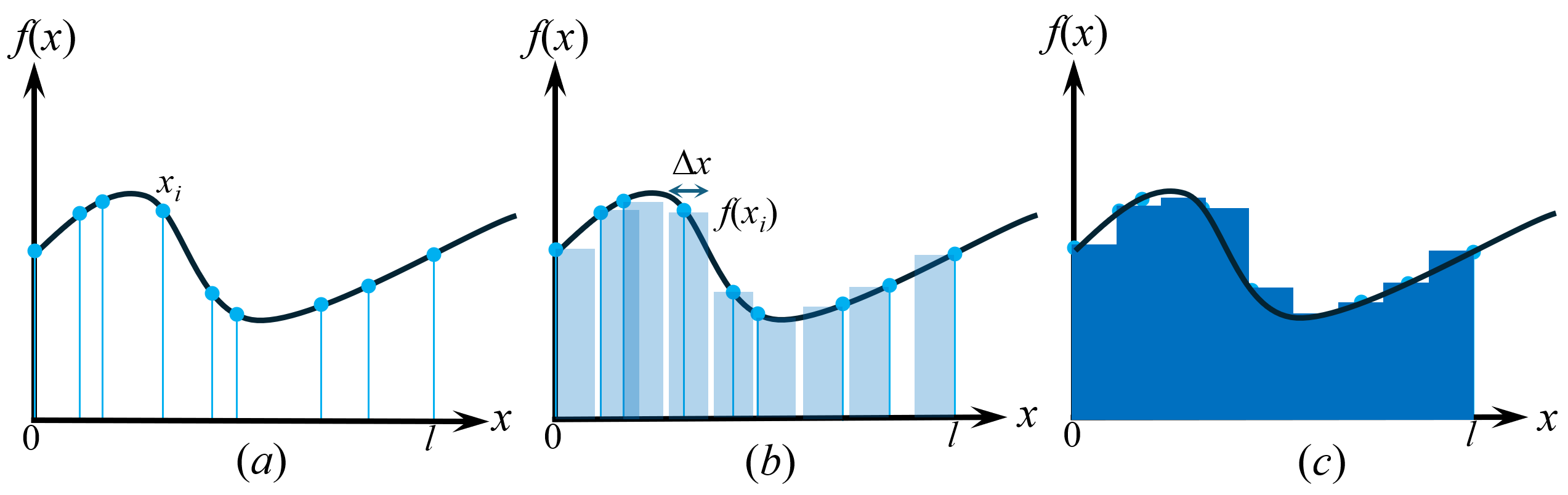}
    \caption{Numerical approximation of a one-dimensional integral. 
    (a) Function \(f(x)\) and sampling points, 
    (b) uniform discretization of the integration domain, and 
    (c) numerical integration result.}
    \label{fig:NI}
\end{figure}

For computational implementation, the continuous integrals in \eqref{eq:KF_def} are approximated by numerical quadrature. Fig.~\ref{fig:NI} illustrates a simple one-dimensional example. For an integral over the interval \([0,l]\), the numerical approximation can be written as
\begin{equation}
    \int_{0}^{l} f(x)\,dx \approx \sum_{i=1}^{N} f(x_i)\Delta x ,
\end{equation}
where \(f(x)\) is a one-dimensional function, \(x_i\) denotes the sampling point, \(N\) is the total number of sampling points, and \(\Delta x=l/N\) is the uniform discretization interval. In Fig.~\ref{fig:NI}(a), the target function and the corresponding sampling points are shown. Fig.~\ref{fig:NI}(b) illustrates the uniform discretization of the integration domain. Fig.~\ref{fig:NI}(c) presents the numerical integration result and the associated approximation error. This numerical integration strategy is used to discretize the weak-form integrals when assembling the matrix system.

Let \(\{\mathbf{x}_i\}_{i=1}^{N_c}\) be the interior integration points sampled in \(\Omega\), and let \(\{\mathbf{x}_j^N\}_{j=1}^{N_N}\) be the boundary integration points sampled on \(\Gamma_N\). The corresponding feature matrix and gradient tensor are denoted by
\begin{equation}
\mathbf{H}
=
\boldsymbol{\Psi}(\mathbf{X},\boldsymbol{\theta}),
\qquad
\mathbf{B}
=
\nabla\boldsymbol{\Psi}(\mathbf{X},\boldsymbol{\theta}),
\end{equation}
where \(\mathbf{H}\in\mathbb{R}^{N_c\times N_h}\), and \(\mathbf{B}\) stores the spatial derivatives of all basis functions at all integration points. Since the proposed basis functions are explicit sinusoidal functions, their gradients can be evaluated analytically. For example, for
\(\psi_j(\mathbf{x})\propto \sin(2\pi\boldsymbol{\omega}_j^T\mathbf{x}+b_j)\), its spatial gradient is given by
\begin{equation}
\nabla\psi_j(\mathbf{x})
\propto
2\pi\boldsymbol{\omega}_j
\cos
\left(
2\pi\boldsymbol{\omega}_j^T\mathbf{x}+b_j
\right).
\end{equation}
Thus, the proposed method avoids the repeated use of automatic differentiation for high-order derivatives and significantly reduces the associated computational overhead.

Using Monte Carlo integration, the stiffness matrix can be approximated as
\begin{equation}
\mathbf{K}
\approx
\frac{|\Omega|}{N_c}
\sum_{i=1}^{N_c}
\left[
\mathbf{B}_i^T
\mathbf{A}(\mathbf{x}_i)
\mathbf{B}_i
+
c(\mathbf{x}_i)
\mathbf{H}_i^T\mathbf{H}_i
\right],
\end{equation}
and the load matrix is approximated by
\begin{equation}
\mathbf{F}
\approx
\frac{|\Omega|}{N_c}
\sum_{i=1}^{N_c}
\mathbf{H}_i^T
\mathbf{f}(\mathbf{x}_i)
+
\frac{|\Gamma_N|}{N_N}
\sum_{j=1}^{N_N}
\mathbf{H}_{N,j}^T
\mathbf{h}(\mathbf{x}_j^N).
\end{equation}
Here, \(|\Omega|\) and \(|\Gamma_N|\) denote the measure of the domain and Neumann boundary, respectively. Other quadrature rules, such as Gaussian quadrature or quasi-Monte Carlo integration, can also be used depending on the geometry and sampling strategy.

In practice, high-dimensional Fourier feature mappings may introduce correlations among basis functions, which can make \(\mathbf{K}\) ill-conditioned. To enhance numerical robustness, Tikhonov regularization is applied to the stiffness matrix:
\begin{equation}
\left(
\mathbf{K}
+
\beta\mathbf{I}
\right)
\mathbf{W}
=
\mathbf{F},
\end{equation}
where \(\beta>0\) is a small regularization parameter and \(\mathbf{I}\) is the identity matrix. The coefficient matrix is then computed as
\begin{equation}
\mathbf{W}
=
\left(
\mathbf{K}
+
\beta\mathbf{I}
\right)^{-1}
\mathbf{F}.
\end{equation}
In numerical implementation, the inverse is not formed explicitly; instead, the linear system is solved using stable direct solvers, such as Cholesky decomposition for symmetric positive-definite systems or QR/SVD-based solvers for ill-conditioned cases. This matrix-learning procedure constitutes the core of the proposed variational matrix-learning Fourier network.

\subsection{Heuristic Scanning Algorithm for Maximum Frequency Selection}
\label{sec:frequency_scanning}
\begin{figure}[!htb]
    \centering
    \includegraphics[width=\linewidth]{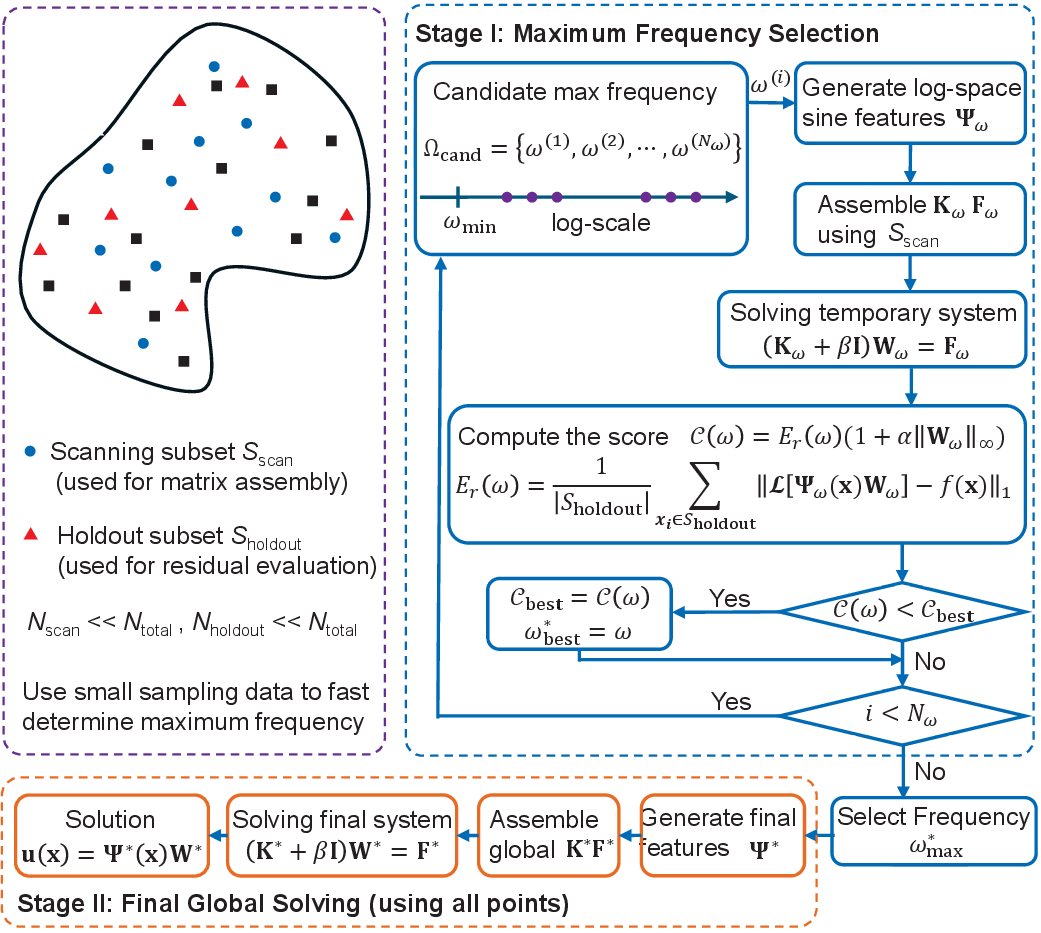}
    \caption{Heuristic scanning algorithm for maximum frequency selection using small sampling points.}
    \label{fig:HFS}
\end{figure}
Although the proposed variational matrix-learning Fourier network (VMLFN) yields a closed-form solution once the approximation space is fixed, its accuracy and numerical stability strongly depend on the prescribed maximum frequency \(\omega_{\max}\) in the log-space Fourier feature construction. If \(\omega_{\max}\) is too small, the generated basis functions contain only low-frequency components, leading to insufficient approximation capability and spectral-bias-induced underfitting. Conversely, an excessively large \(\omega_{\max}\) introduces highly oscillatory basis functions, which may amplify numerical quadrature errors, deteriorate the conditioning of the stiffness matrix, and cause large output coefficients. These effects may eventually produce spurious oscillations and unstable predictions. Therefore, an appropriate choice of \(\omega_{\max}\) is essential for balancing expressiveness and stability.

To avoid manual hyperparameter tuning, this work proposes a physics-informed heuristic frequency scanning algorithm for automatic selection of \(\omega_{\max}\), as shown in Fig.~\ref{fig:HFS}. Unlike conventional data-driven model selection strategies that require labeled validation data, the proposed criterion evaluates each candidate frequency bound using only the governing physics and the numerical stability of the resulting matrix system. The key idea is to construct a low-cost temporary variational solve for each candidate frequency and assess its physical consistency on an unseen set of collocation points.

Let \(\boldsymbol{\Omega}_{cand}=\{\omega^{(1)},\omega^{(2)},\ldots,\omega^{(N_\omega)}\}\) denote a prescribed set of candidate maximum frequencies. The full set of collocation points is randomly divided into two disjoint subsets: a small scanning subset \(\mathcal{S}_{scan}\) and a holdout evaluation subset \(\mathcal{S}_{holdout}\). The scanning subset is used to assemble a temporary stiffness matrix and load vector, while the holdout subset is used only for evaluating the physical residual. In this way, the selected frequency is encouraged to generalize to unseen collocation points rather than merely minimizing the residual on the points used for matrix assembly.

For each candidate \(\omega_{\max}\), a temporary basis \(\boldsymbol{\Psi}_{\omega}\) is generated according to the log-space sampling rule bounded by \(\omega_{\max}\). The corresponding temporary coefficient matrix is obtained by solving
\begin{equation}
\mathbf{W}_{\omega}
=
\left(
\mathbf{K}_{\omega}
+
\beta\mathbf{I}
\right)^{-1}
\mathbf{F}_{\omega},
\label{eq:tmp_solve}
\end{equation}
where \(\mathbf{K}_{\omega}\) and \(\mathbf{F}_{\omega}\) are assembled using only the scanning subset \(\mathcal{S}_{scan}\), and \(\beta\) is the Tikhonov regularization parameter. The physical residual is then evaluated on the holdout subset as
\begin{equation}
\mathbf{r}_{\omega}(\mathbf{x})
=
\mathcal{L}
\left[
\boldsymbol{\Psi}_{\omega}(\mathbf{x})\mathbf{W}_{\omega}
\right]
-
\mathbf{f}(\mathbf{x}),
\qquad
\mathbf{x}\in\mathcal{S}_{holdout},
\end{equation}
where \(\mathcal{L}[\cdot]\) denotes the differential operator of the governing PDE. The holdout residual error is measured by
\begin{equation}
E_{r}(\omega_{\max})
=
\frac{1}{|\mathcal{S}_{holdout}|}
\sum_{\mathbf{x}_i\in\mathcal{S}_{holdout}}
\left\|
\mathbf{r}_{\omega}(\mathbf{x}_i)
\right\|_1 .
\end{equation}

To jointly account for physical accuracy and numerical stability, the following frequency-selection score is introduced:
\begin{equation}
\mathcal{C}(\omega_{\max})
=
E_{r}(\omega_{\max})
\left(
1
+
\alpha
\left\|
\mathbf{W}_{\omega}
\right\|_{\infty}
\right),
\label{eq:frequency_score}
\end{equation}
where \(\alpha\) is a small positive penalty coefficient. In this work, \(\alpha=10^{-3}\) is used unless otherwise specified. The first term in \eqref{eq:frequency_score} measures the physical consistency of the temporary solution on unseen collocation points, whereas the second term penalizes excessively large output weights. This penalty is important because overly high-frequency bases often induce near-linear dependence among features and degrade the conditioning of \(\mathbf{K}_{\omega}\), leading to an abnormal increase in \(\|\mathbf{W}_{\omega}\|_{\infty}\). Therefore, the proposed score acts as a stability-aware residual metric and suppresses pathological frequency choices.

The optimal maximum frequency is selected by
\begin{equation}
\omega_{\max}^{*}
=
\arg\min_{\omega_{\max}\in\boldsymbol{\Omega}_{cand}}
\mathcal{C}(\omega_{\max}).
\label{eq:omega_opt}
\end{equation}
After \(\omega_{\max}^{*}\) is determined, the final basis functions are regenerated using the selected frequency bound, and the global variational matrix system is assembled using the full set of integration points. The final coefficient matrix is then computed by a regularized linear solve.

The complete auto-tuning and matrix-learning workflow of VMLFN is summarized in Algorithm~\ref{alg:vmlfn_scan}. Since the scanning phase uses only a small number of collocation points and relies on analytical derivatives of the sine features, the computational overhead is negligible compared with the final global assembly and solution stage.

\begin{algorithm}[!t]
\caption{VMLFN With Heuristic Frequency Scanning}
\label{alg:vmlfn_scan}
\begin{algorithmic}[1]
\Require Computational domain \(\Omega\), governing PDE operator \(\mathcal{L}\), source term \(\mathbf{f}\), scanning point budget \(N_{scan}\), holdout point budget \(N_{holdout}\), candidate frequency set \(\boldsymbol{\Omega}_{cand}=\{\omega^{(1)},\omega^{(2)},\ldots,\omega^{(N_\omega)}\}\), regularization parameter \(\beta\), penalty coefficient \(\alpha\).
\Ensure Selected maximum frequency \(\omega_{\max}^{*}\), coefficient matrix \(\mathbf{W}^{*}\), predicted field \(\mathbf{u}(\mathbf{x})\).

\State Randomly sample the scanning subset \(\mathcal{S}_{scan}\subset\Omega\) with \(N_{scan}\) points.
\State Randomly sample the holdout subset \(\mathcal{S}_{holdout}\subset\Omega\) with \(N_{holdout}\) points.
\State Initialize \(C_{best}\leftarrow \infty\) and \(\omega_{\max}^{*}\leftarrow \omega^{(1)}\).

\For{each candidate \(\omega\in\boldsymbol{\Omega}_{cand}\)}
    \State Generate temporary log-space sine features \(\boldsymbol{\Psi}_{\omega}\) with maximum frequency \(\omega\).
    \State Evaluate \(\boldsymbol{\Psi}_{\omega}\) and analytical gradients \(\nabla\boldsymbol{\Psi}_{\omega}\) on \(\mathcal{S}_{scan}\).
    \State Assemble temporary matrices \(\mathbf{K}_{\omega}\) and \(\mathbf{F}_{\omega}\) using \(\mathcal{S}_{scan}\).
    \State Solve the regularized temporary system:
    \[
    \mathbf{W}_{\omega}
    =
    \left(
    \mathbf{K}_{\omega}
    +
    \beta\mathbf{I}
    \right)^{-1}
    \mathbf{F}_{\omega}.
    \]
    \State Evaluate the PDE residual \(\mathbf{r}_{\omega}\) on \(\mathcal{S}_{holdout}\).
    \State Compute the holdout residual error:
    \[
    E_r(\omega)
    =
    \frac{1}{|\mathcal{S}_{holdout}|}
    \sum_{\mathbf{x}_i\in\mathcal{S}_{holdout}}
    \left\|
    \mathbf{r}_{\omega}(\mathbf{x}_i)
    \right\|_1 .
    \]
    \State Compute the stability-aware score:
    \[
    \mathcal{C}(\omega)
    =
    E_r(\omega)
    \left(
    1+\alpha\|\mathbf{W}_{\omega}\|_{\infty}
    \right).
    \]
    \If{\(\mathcal{C}(\omega)<C_{best}\)}
        \State \(C_{best}\leftarrow \mathcal{C}(\omega)\).
        \State \(\omega_{\max}^{*}\leftarrow \omega\).
    \EndIf
\EndFor

\State Generate the final log-space sine features \(\boldsymbol{\Psi}^{*}\) using \(\omega_{\max}^{*}\).
\State Evaluate \(\boldsymbol{\Psi}^{*}\) and \(\nabla\boldsymbol{\Psi}^{*}\) on the full integration set.
\State Assemble the global stiffness matrix \(\mathbf{K}^{*}\) and load matrix \(\mathbf{F}^{*}\).
\State Solve the final regularized matrix system:
\[
\mathbf{W}^{*}
=
\left(
\mathbf{K}^{*}
+
\beta\mathbf{I}
\right)^{-1}
\mathbf{F}^{*}.
\]
\State Reconstruct the predicted solution:
\[
\mathbf{u}(\mathbf{x})
=
\boldsymbol{\Psi}^{*}(\mathbf{x})\mathbf{W}^{*}.
\]
\State \Return \(\omega_{\max}^{*}\), \(\mathbf{W}^{*}\), and \(\mathbf{u}(\mathbf{x})\).
\end{algorithmic}
\label{alg:HFS}
\end{algorithm}

This two-stage strategy equips VMLFN with an automatic frequency-selection capability. By screening candidate approximation spaces through a physics-informed holdout residual and a stability-aware weight penalty, the method adaptively selects a frequency bandwidth that is compatible with the spectral complexity of the target physical problem. Consequently, the final variational matrix solve can achieve improved accuracy while maintaining numerical robustness.

\subsection{Specific Physical Field Instantiations}
\label{subsec:physical_instantiations}

Based on the proposed variational matrix-learning Fourier network (VMLFN), different engineering physics problems can be instantiated in a unified manner by specifying the material tensor \(\mathbf{A}\), reaction coefficient \(c\), source term \(\mathbf{f}\), and boundary flux \(\mathbf{h}\). This subsection presents several representative cases, including scalar Helmholtz and heat conduction problems, parametric surrogate modeling, and vector-valued thermoelastic deformation. These examples demonstrate that the proposed VMLFN can be directly adapted to different governing equations without changing the core matrix-learning procedure.

\subsubsection{Case I: 3-D Active Helmholtz Equation}

The active Helmholtz equation is commonly used to describe steady-state acoustic and electromagnetic wave propagation. Its scalar form is given by
\begin{equation}
-\nabla^2 u - k^2 u = f(\mathbf{x}),
\qquad
\mathbf{x}\in\Omega ,
\end{equation}
where \(u\) is the wave field, \(k\) is the wavenumber, and \(f(\mathbf{x})\) denotes the source term. In this case, a homogeneous Dirichlet boundary condition is imposed on the bottom surface \(z=0\), denoted by \(\Gamma_D\), and homogeneous Neumann conditions are imposed on the remaining boundaries \(\Gamma_N\).

By mapping the Helmholtz equation to the unified second-order form, one has
\begin{equation}
\mathbf{A}=\mathbf{I},
\qquad
c=-k^2,
\qquad
\mathbf{h}=0 .
\end{equation}
Accordingly, the energy functional can be written as
\begin{equation}
J(u)
=
\frac{1}{2}
\int_{\Omega}
\left(
\nabla u\cdot\nabla u
-
k^2u^2
\right)
d\Omega
-
\int_{\Omega}
fu
\,d\Omega .
\label{eq:helmholtz_energy}
\end{equation}
To enforce the homogeneous Dirichlet condition \(u=0\) on the bottom surface, the spatial coordinates are first normalized to \([-1,1]^3\), and a distance-type envelope function is introduced:
\begin{equation}
\mathbf{D}(\mathbf{x})
=
\tanh
\left[
\gamma
\left(
z_{\rm norm}+1
\right)
\right],
\end{equation}
where \(\gamma>0\) controls the transition rate. The VMLFN trial solution is then constructed as
\begin{equation}
\mathbf{u}(\mathbf{x})
=
\boldsymbol{\Psi}(\mathbf{x})
\mathbf{W}.
\end{equation}
Let \(\mathbf{H}\) be the evaluation matrix of the modulated basis \(\boldsymbol{\Psi}\), and let \(\mathbf{B}_x\), \(\mathbf{B}_y\), and \(\mathbf{B}_z\) denote its analytical spatial derivative matrices. The corresponding stiffness matrix and load vector are assembled as
\begin{equation}
\begin{split}
\mathbf{K}_{H}
&=
\mathbf{B}_x^T\mathbf{B}_x
+
\mathbf{B}_y^T\mathbf{B}_y
+
\mathbf{B}_z^T\mathbf{B}_z
-
k^2\mathbf{H}^T\mathbf{H}, \\
\mathbf{F}_{H}
&=
\mathbf{H}^T\mathbf{f}.
\end{split}
\label{eq:helmholtz_KF}
\end{equation}
In practice, the matrices in \eqref{eq:helmholtz_KF} are multiplied by the corresponding quadrature or Monte Carlo integration weights.

\subsubsection{Case II: 3-D Steady-State Heat Conduction With Homogeneous Neumann Boundaries}

For a homogeneous medium, the steady-state heat conduction equation is given by
\begin{equation}
-\kappa\nabla^2 T = Q(\mathbf{x}),
\qquad
\mathbf{x}\in\Omega ,
\end{equation}
where \(T\) is the temperature field, \(\lambda\) is the thermal conductivity, and \(Q(\mathbf{x})\) is the volumetric heat source. The bottom surface is maintained at a prescribed temperature \(T_{\rm ref}\), while the remaining surfaces are assumed to be adiabatic.

The corresponding coefficients in the unified framework are
\begin{equation}
\mathbf{A}=\kappa\mathbf{I},
\qquad
c=0,
\qquad
\mathbf{h}=0 .
\end{equation}
The thermal energy functional is therefore
\begin{equation}
J(T)
=
\frac{1}{2}
\int_{\Omega}
\kappa
\nabla T\cdot\nabla T
\,d\Omega
-
\int_{\Omega}
QT
\,d\Omega .
\label{eq:heat_energy}
\end{equation}
To impose the nonhomogeneous Dirichlet boundary condition \(T=T_{\rm ref}\) on the bottom surface, the trial solution is defined as
\begin{equation}
T_h(\mathbf{x})
=
T_{\rm ref}
+
\boldsymbol{\Psi}(\mathbf{x})
\mathbf{W}.
\end{equation}
Since \(T_{\rm ref}\) is a constant, its gradient vanishes. Thus, the gradient of the trial solution is determined only by the modulated Fourier basis. Let \(\mathbf{H}\) denote the modulated basis evaluation matrix and \(\mathbf{B}_x\), \(\mathbf{B}_y\), and \(\mathbf{B}_z\) denote its spatial derivative matrices. The thermal stiffness matrix and load vector become
\begin{equation}
\begin{split}
\mathbf{K}_{T}
&=
\kappa
\left(
\mathbf{B}_x^T\mathbf{B}_x
+
\mathbf{B}_y^T\mathbf{B}_y
+
\mathbf{B}_z^T\mathbf{B}_z
\right), \\
\mathbf{F}_{T}
&=
\mathbf{H}^T\mathbf{Q}.
\end{split}
\label{eq:heat_KF}
\end{equation}

\subsubsection{Case III: Heat Conduction With a Nonhomogeneous Neumann Boundary}

In practical thermal-management problems, boundary surfaces are often subjected to prescribed heat fluxes rather than ideal adiabatic conditions. Consider the case where a heat flux \(q_{\rm top}\) is applied on the top surface \(\Gamma_{\rm top}\), while the other non-Dirichlet surfaces remain adiabatic. The Neumann condition is written as
\begin{equation}
\kappa\nabla T\cdot\mathbf{n}
=
q_{\rm top},
\qquad
\mathbf{x}\in\Gamma_{\rm top}.
\end{equation}
In the VMLFN formulation, this nonhomogeneous Neumann boundary condition is incorporated naturally through the boundary integral in the variational functional:
\begin{equation}
\begin{split}
J(T)
=
&
\frac{1}{2}
\int_{\Omega}
\kappa
\nabla T\cdot\nabla T
\,d\Omega
-
\int_{\Omega}
QT
\,d\Omega  \\
&
-
\int_{\Gamma_{\rm top}}
q_{\rm top}T
\,d\Gamma .
\end{split}
\label{eq:heat_flux_energy}
\end{equation}
Therefore, the stiffness matrix \(\mathbf{K}_{T}\) remains identical to that in \eqref{eq:heat_KF}. Only the load vector is modified by adding the projected boundary-flux contribution:
\begin{equation}
\mathbf{F}_{T}
=
\mathbf{H}^T\mathbf{Q}
+
\mathbf{H}_{\rm top}^T\mathbf{q}_{\rm top},
\label{eq:heat_flux_load}
\end{equation}
where \(\mathbf{H}_{\rm top}\) is the basis evaluation matrix on \(\Gamma_{\rm top}\), and \(\mathbf{q}_{\rm top}\) is the sampled heat-flux vector. The volume and surface integration weights are omitted in \eqref{eq:heat_flux_load} for compactness but are included in the numerical implementation. This case illustrates a key advantage of the variational formulation: prescribed fluxes are handled as natural boundary conditions without introducing additional penalty weights.

\subsubsection{Case IV: 4-D Parametric Surrogate Modeling for Variable Conductivity}

For engineering design and optimization, the PDE often needs to be solved repeatedly under different material properties. To avoid repeated simulations, VMLFN can be extended to a spatio-parametric surrogate model by augmenting the input coordinates with the material parameter. For the variable-conductivity heat conduction problem, the augmented input is defined as
\begin{equation}
\mathbf{s}
=
[x,y,z,\kappa]^T
\in
\Omega\times\Lambda ,
\qquad
\Lambda=
[\kappa_{\min},\kappa_{\max}] .
\end{equation}
The Fourier basis is therefore constructed as
\begin{equation}
\boldsymbol{\Psi}
=
\boldsymbol{\Psi}(x,y,z,\kappa),
\end{equation}
where the frequency vectors are sampled in the four-dimensional augmented input space. It is important to note that the physical differential operator acts only on the spatial variables. Therefore, the gradient used for matrix assembly is
\begin{equation}
\nabla_{\rm s}
\boldsymbol{\Psi}
=
\left[
\partial_x\boldsymbol{\Psi},
\partial_y\boldsymbol{\Psi},
\partial_z\boldsymbol{\Psi}
\right]^T,
\end{equation}
where derivatives with respect to \(\lambda\) are not included in the PDE operator.

By sampling both spatial coordinates and conductivity values, the variational matrices are assembled over the augmented domain \(\Omega\times\Lambda\). Let \(\boldsymbol{\kappa}_{\rm col}\) be the vector of sampled conductivity values at collocation points. The stiffness matrix of the 4-D surrogate model is assembled as
\begin{equation}
\begin{split}
\mathbf{K}_{4D}
=
\frac{1}{N_{\rm total}}
\sum
\Big[
&
\mathbf{B}_x^T
\left(
\boldsymbol{\kappa}_{\rm col}
\odot
\mathbf{B}_x
\right)
+
\mathbf{B}_y^T
\left(
\boldsymbol{\kappa}_{\rm col}
\odot
\mathbf{B}_y
\right)  \\
&
+
\mathbf{B}_z^T
\left(
\boldsymbol{\kappa}_{\rm col}
\odot
\mathbf{B}_z
\right)
\Big],
\end{split}
\label{eq:4D_K}
\end{equation}
where \(\odot\) denotes row-wise Hadamard multiplication. The corresponding load vector is assembled using the sampled source term \(Q(\mathbf{x},\kappa)\). Once the coefficient matrix is solved, the trained VMLFN model can predict the thermal field for any \(\kappa\in\Lambda\) without reassembling a new PDE system.

\subsubsection{Case V: 3-D Thermoelastic Warping in Heterogeneous Composites}

The final example considers a vector-valued thermoelastic deformation problem in heterogeneous composites. In strong form, the Navier--Cauchy equation with spatially varying elastic parameters involves derivatives of the Lamé constants and second-order derivatives of the displacement field. This may lead to complicated differential operators and expensive automatic differentiation when solved by strong-form neural methods. In contrast, the variational formulation only requires the first-order strain tensor and is therefore more suitable for heterogeneous elasticity.

Let the displacement field be
\begin{equation}
\mathbf{u}
=
[u_x,u_y,u_z]^T .
\end{equation}
The small-strain tensor is defined as
\begin{equation}
\boldsymbol{\varepsilon}
=
\frac{1}{2}
\left(
\nabla\mathbf{u}
+
\nabla\mathbf{u}^T
\right),
\end{equation}
and the thermoelastic potential energy is written as
\begin{equation}
\Pi(\mathbf{u})
=
\frac{1}{2}
\int_{\Omega}
\boldsymbol{\varepsilon}^T
\mathbf{C}
\boldsymbol{\varepsilon}
\,d\Omega
-
\int_{\Omega}
\boldsymbol{\varepsilon}^T
\mathbf{C}
\boldsymbol{\varepsilon}_{th}
\,d\Omega ,
\label{eq:elastic_energy}
\end{equation}
where \(\mathbf{C}\) is the elasticity matrix and \(\boldsymbol{\varepsilon}_{th}\) is the thermal strain vector. For isotropic thermoelasticity,
\begin{equation}
\boldsymbol{\varepsilon}_{th}
=
\alpha\Delta T
[1,1,1,0,0,0]^T ,
\end{equation}
where \(\alpha\) is the thermal expansion coefficient and \(\Delta T\) is the temperature change.

Assuming traction-free boundaries, no Dirichlet envelope is required in this case. The three displacement components are approximated by the same Fourier basis:
\begin{equation}
u_x=\boldsymbol{\Psi}\mathbf{W}_x,
\qquad
u_y=\boldsymbol{\Psi}\mathbf{W}_y,
\qquad
u_z=\boldsymbol{\Psi}\mathbf{W}_z .
\end{equation}
Equivalently, the unknown coefficients are stacked as
\begin{equation}
\mathbf{W}
=
\left[
\mathbf{W}_x^T,
\mathbf{W}_y^T,
\mathbf{W}_z^T
\right]^T
\in
\mathbb{R}^{3N_h\times 1}.
\end{equation}

For heterogeneous composites, the Lamé parameters \(\lambda_e(\mathbf{x})\) and \(\mu_e(\mathbf{x})\) are evaluated at the collocation points and stored as vectors \(\boldsymbol{\lambda}_e\) and \(\boldsymbol{\mu}_e\). The global elastic stiffness matrix
\(\mathbf{K}_E\in\mathbb{R}^{3N_h\times 3N_h}\) is assembled using \(3\times3\) block matrices. For example, the diagonal \(u_x\)-\(u_x\) block and the coupling \(u_x\)-\(u_y\) block are
\begin{equation}
\begin{split}
\mathbf{K}_{00}
=&
\mathbf{B}_x^T
\left[
(\boldsymbol{\lambda}_e+2\boldsymbol{\mu}_e)
\odot
\mathbf{B}_x
\right]
+
\mathbf{B}_y^T
\left[
\boldsymbol{\mu}_e
\odot
\mathbf{B}_y
\right]  \\
&+
\mathbf{B}_z^T
\left[
\boldsymbol{\mu}_e
\odot
\mathbf{B}_z
\right],
\end{split}
\label{eq:elastic_K00}
\end{equation}
and
\begin{equation}
\begin{split}
\mathbf{K}_{01}
=
\mathbf{K}_{10}^T
=&
\mathbf{B}_x^T
\left(
\boldsymbol{\lambda}_e
\odot
\mathbf{B}_y
\right)
+
\mathbf{B}_y^T
\left(
\boldsymbol{\mu}_e
\odot
\mathbf{B}_x
\right).
\end{split}
\label{eq:elastic_K01}
\end{equation}
The remaining block entries are obtained in an analogous manner by cyclic permutation of the spatial derivatives. The thermal load vector associated with the \(x\)-displacement component is assembled as
\begin{equation}
\mathbf{F}_{E_x}
=
\mathbf{B}_x^T
\left[
(3\boldsymbol{\lambda}_e+2\boldsymbol{\mu}_e)
\odot
\boldsymbol{\alpha}
\odot
\Delta\mathbf{T}
\right],
\label{eq:elastic_Fx}
\end{equation}
with similar expressions for \(\mathbf{F}_{E_y}\) and \(\mathbf{F}_{E_z}\) using \(\mathbf{B}_y^T\) and \(\mathbf{B}_z^T\), respectively. Solving the resulting block linear system gives the three-dimensional thermoelastic warping field in a single matrix-learning step.

These instantiations show that VMLFN provides a unified and extensible framework for scalar, parametric, and vector-valued physical fields. By changing only the coefficients, load terms, and boundary contributions in the variational matrices, the same Fourier feature representation and closed-form matrix-learning solver can be reused across different classes of engineering PDEs.

\section{Numerical Experiments and Results}
\label{sec:experimentalresults}
In this section, five representative benchmark problems are considered to comprehensively evaluate the robustness, accuracy, and computational efficiency of the proposed variational matrix-learning Fourier network (VMLFN). The test cases cover scalar elliptic equations, heat conduction with different boundary conditions, parametric surrogate modeling, and vector-valued thermoelastic deformation. The numerical results are systematically compared with those obtained from a conventional physics-informed neural network (PINN) and finite-element solutions computed by COMSOL Multiphysics.

All experiments are conducted in a CUDA-enabled PyTorch environment on an NVIDIA GPU. Depending on the conditioning of the resulting linear system and the complexity of the target field, either full double-precision floating-point arithmetic (FP64) or single-precision arithmetic (FP32) is employed. 

For a rigorous and fair comparison, the baseline PINN is implemented using a standard multilayer perceptron (MLP) with four hidden layers and 128 neurons per layer. The hyperbolic tangent function is adopted as the activation function. The PINN parameters are optimized using the Adam optimizer with a learning-rate scheduler. The total loss function is defined as
\begin{equation}
\mathcal{L}_{\rm Total}
=
\mathcal{L}_{\rm PDE}
+
\omega_{\rm Dir}\mathcal{L}_{\rm Dir}
+
\omega_{\rm Neu}\mathcal{L}_{\rm Neu},
\label{eq:pinn_loss}
\end{equation}
where \(\mathcal{L}_{\rm PDE}\), \(\mathcal{L}_{\rm Dir}\), and \(\mathcal{L}_{\rm Neu}\) denote the residual loss, Dirichlet boundary loss, and Neumann boundary loss, respectively. The boundary penalty weights are empirically tuned for stable convergence. In the present implementation, typical values are set to \(\omega_{\rm Dir}=50.0\) and \(\omega_{\rm Neu}=10.0\), unless further adjustment is required for a specific case.

In contrast, VMLFN avoids iterative nonlinear training by converting the variational problem into a matrix-learning problem. Dirichlet boundary conditions are imposed through an analytical envelope function, while Neumann boundary conditions are naturally incorporated through the boundary integral of the variational functional. Therefore, no manually tuned boundary penalty weights are required. The hidden dimension of VMLFN is selected in the range
$N_h = 64 \sim 8000$,
depending on the complexity and frequency content of the target physical field. 

\begin{figure*}[!htb]
    \centering
    \subfigure[]{\includegraphics[width=0.2\textwidth]{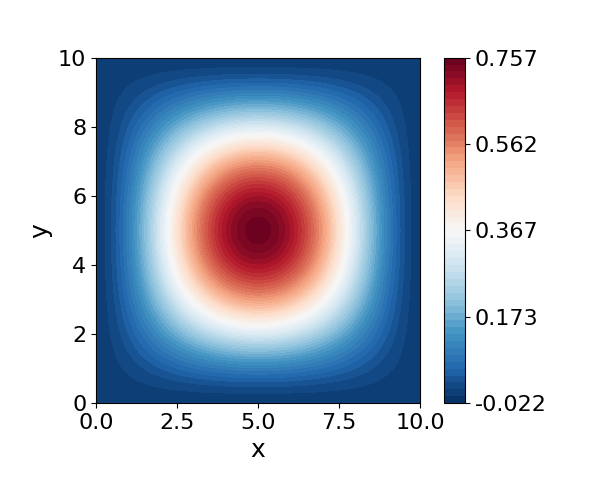}}%
    \subfigure[]{\includegraphics[width=0.2\textwidth]{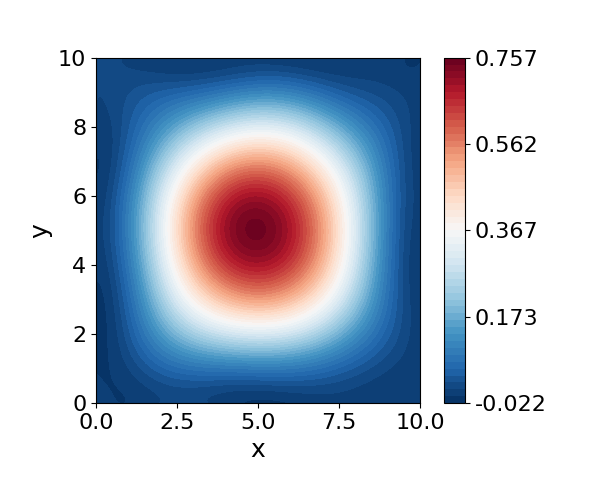}}%
    \subfigure[]{\includegraphics[width=0.2\textwidth]{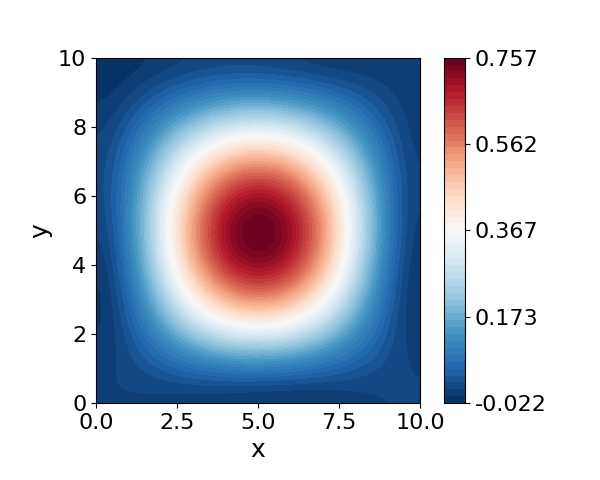}}%
    \subfigure[]{\includegraphics[width=0.2\textwidth]{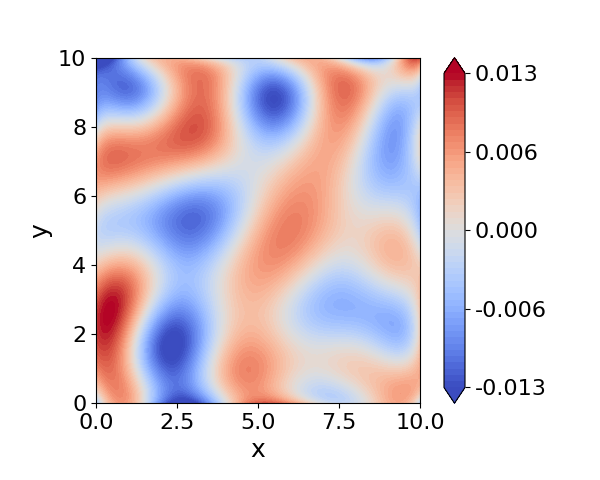}}%
    \subfigure[]{\includegraphics[width=0.2\textwidth]{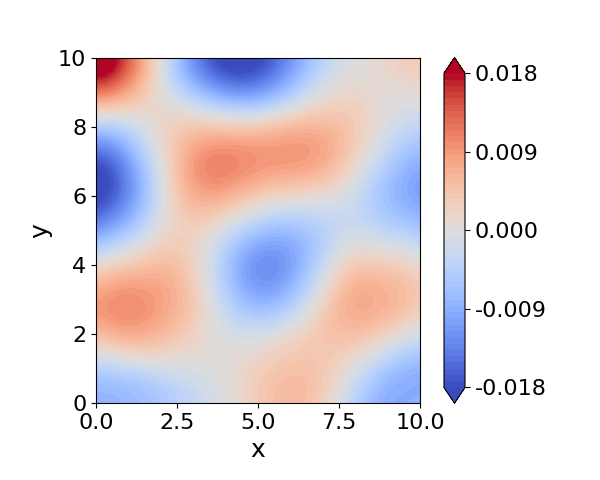}}
    \caption{Wave-field distributions for the original polynomial manufactured solution on the representative slice \(z=2\). (a) Exact solution, (b) VMLFN prediction, (c) PINN prediction, (d) absolute error of VMLFN, and (e) absolute error of PINN. }
    \label{fig:helmholtz_poly_visual}
\end{figure*}

\begin{figure*}[!htb]
    \centering
    \subfigure[]{\includegraphics[width=0.2\textwidth]{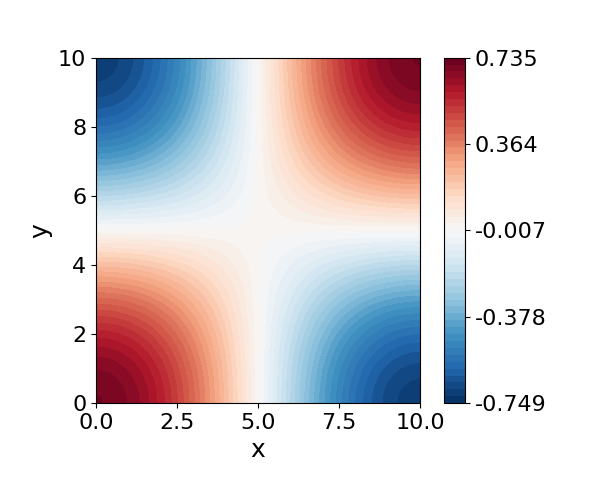}}%
    \subfigure[]{\includegraphics[width=0.2\textwidth]{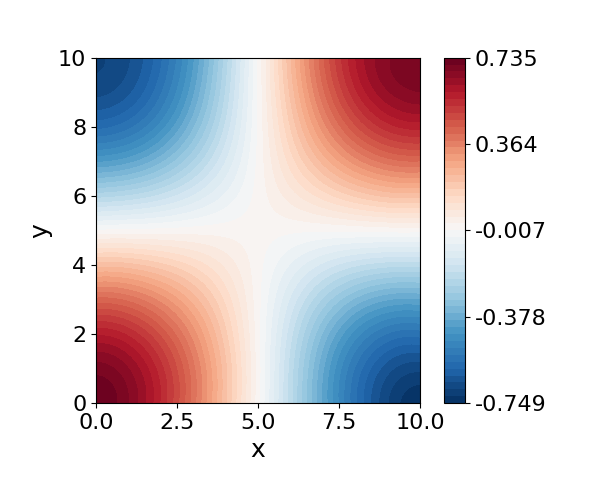}}%
    \subfigure[]{\includegraphics[width=0.2\textwidth]{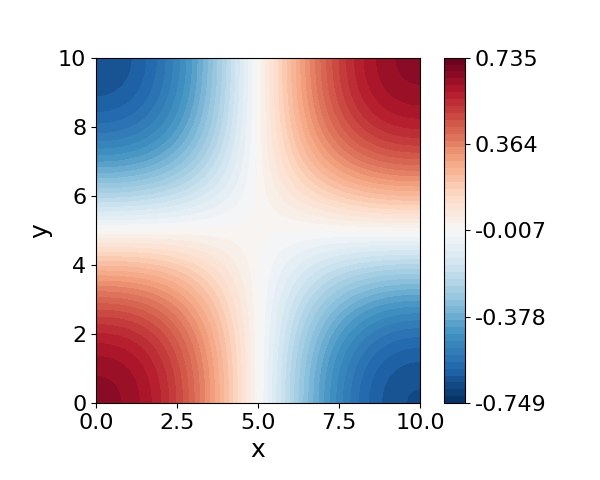}}%
    \subfigure[]{\includegraphics[width=0.2\textwidth]{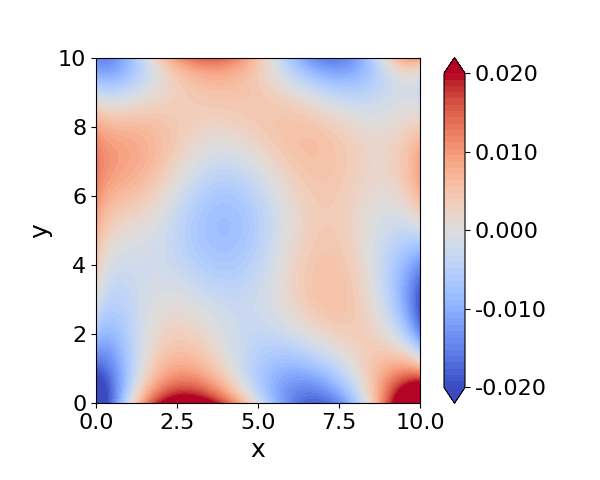}}%
    \subfigure[]{\includegraphics[width=0.2\textwidth]{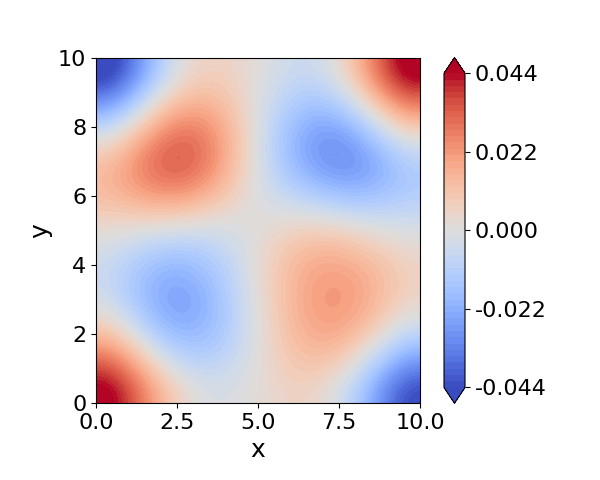}}
    \caption{Wave-field distributions for the high-frequency cosine--cosine--sine manufactured solution on the representative slice \(z=2\). (a) Exact solution, (b) VMLFN prediction, (c) PINN prediction, (d) absolute error of VMLFN, and (e) absolute error of PINN.}
    \label{fig:helmholtz_cos_visual}
\end{figure*}

The performance of all methods is evaluated using two metrics. 
The first metric is the average absolute field difference with respect to the analytical, or COMSOL reference solution:
\begin{equation}
{\rm Avg.\ Diff.}
=
\frac{1}{N_t}
\sum_{i=1}^{N_t}
\left|
\hat{s}
\left(
\mathbf{x}_i
\right)
-
s_{\rm ref}
\left(
\mathbf{x}_i
\right)
\right|,
\label{eq:avg_diff}
\end{equation}
where \(\hat{s}\) denotes the predicted temperature, displacement, or scalar solution, and \(s_{\rm ref}\) denotes the corresponding reference field. For vector-valued displacement problems, the absolute difference is computed componentwise and then averaged over all displacement components and test points. The second metric is the total execution time, including matrix assembly and solution for VMLFN, and the complete optimization time for PINN.

These metrics provide a practical assessment of both predictive accuracy and computational efficiency. Specifically, the average field difference quantifies the discrepancy between the predicted solution and the analytical or COMSOL reference field, while the total execution time reflects the overall computational cost of each method. For vector-valued displacement problems, the absolute difference is computed componentwise and then averaged over all displacement components and test points. For VMLFN, the reported time includes matrix assembly and solution, whereas for PINN it corresponds to the complete optimization time.

\begin{table}
\centering
\caption{Quantitative Performance Comparison on the 3D Active Helmholtz Equation across two manufactured solutions. Evaluated on an unseen testing grid of $N_{val} = 2000$ points.}
\label{tab:helmholtz_results}
\resizebox{\linewidth}{!}{\begin{tabular}{lcccccc}
\toprule
\textbf{Equation Form} & \textbf{Method} & \textbf{Params}  & \textbf{Rel. Error} & \textbf{Time (s)} & \textbf{Speedup} \\ \midrule
\multirow{2}{*}{Polynomial} & PINN & $50,177$ &  $2.18\%$ & $623.73$ & $1\times$ \\
& VMLFN & $\mathbf{2,000}$ & $\mathbf{1.98\%}$ & $\mathbf{0.17}$ & $\mathbf{3669\times}$ \\ \midrule
\multirow{2}{*}{Cos-Cos-Sin} & PINN & $50,177$  & $1.63\%$ & $609.14$ & $1\times$ \\
& VMLFN & $\mathbf{2,000}$  & $1.75\%$ & $\mathbf{0.06}$ & $\mathbf{>10,000\times}$ \\ 
\bottomrule
\multicolumn{6}{l}{$\omega_{\mathrm{max}}^* = 1.79$ (Polynomial) and $\omega_{\mathrm{max}}^* = 0.43$ (Cos-Cos-Sin).} \\
\end{tabular}}
\end{table}

    



\subsection{Case I: 3-D Active Helmholtz Equation}
\label{subsec:case_helmholtz}

For the 3-D active Helmholtz equation, the main numerical challenge is to accurately capture oscillatory wave fields while alleviating the spectral bias commonly observed in standard neural-network approximators. The computational domain is defined as $\Omega=[0,10]\times[0,10]\times[0,2]$, with the wavenumber set to \(k=1.4\). To examine the robustness of the proposed solver, two manufactured analytical solutions are considered: an original polynomial manufactured solution (Polynomial MMS) and a high-frequency cosine--cosine--sine manufactured solution (Cos-Cos-Sin MMS). All experiments in this case are performed using single-precision floating-point arithmetic (FP32) with a fixed random seed of 2026.

\subsubsection{Model Configuration}

The proposed variational matrix-learning Fourier network is configured with \(N_h=2000\) Fourier basis functions and no additional bias term. Consequently, the model contains only 2000 trainable output weights, and the resulting variational problem reduces to a linear algebraic system. During the automatic bandwidth-tuning stage, the upper bound of the random Fourier frequency, denoted by \(\omega_{\max}\), is selected through a heuristic scan. For the Polynomial MMS, \(\omega_{\max}\) is scanned within the interval $\omega_{\max}\in[0.01,4.0]$
using 15 uniformly spaced candidates. For the Cos-Cos-Sin MMS, the scan is performed over
$\omega_{\max}\in[0.01,0.43]$
using 5 uniformly spaced candidates. To reduce the tuning overhead, each candidate bandwidth is evaluated using only 800 collocation points. After the optimal bandwidth is selected, the final global matrix is assembled using \(N_{\rm train}=8000\) interior collocation points. The linear system is stabilized using Tikhonov regularization with \(\beta=10^{-4}\), and the steepness parameter of the Dirichlet envelope is set to \(\gamma=5.0\).

For comparison, the conventional PINN baseline adopts a fully connected MLP with four hidden layers and 128 neurons per hidden layer. The hyperbolic tangent function is used as the activation function, leading to 50,177 trainable parameters. The PINN is trained for 60,000 epochs using the Adam optimizer with an initial learning rate of \(10^{-3}\). Mini-batches of 1024 interior points are used for the PDE residual, and mini-batches of 512 points are used for each boundary condition. For boundary sampling, 600 uniformly distributed points are generated on each face of the domain. The boundary penalty weights are fixed as $\omega_{\rm Dir}=50.0$ and $\omega_{\rm Neu}=10.0$.

\subsubsection{Quantitative and Visual Comparison}

As summarized in Table~\ref{tab:helmholtz_results}, the conventional PINN requires more than 10 minutes of iterative gradient-based training, with a total execution time of approximately \(610\)--\(623\) s for the two manufactured solutions. In contrast, VMLFN avoids nonlinear optimization and obtains the output weights through direct matrix assembly and closed-form solution. For the Polynomial MMS and the Cos-Cos-Sin MMS, VMLFN achieves relative errors of \(1.98\%\) and \(1.75\%\), respectively. The corresponding total computational times are only \(0.17\) s and \(0.06\) s.

These results indicate that VMLFN attains competitive accuracy with substantially fewer trainable parameters. Specifically, compared with the 50,177-parameter PINN, the proposed method uses only 2000 output weights, corresponding to approximately a \(25\times\) reduction in trainable parameters. More importantly, the elimination of iterative backpropagation leads to a computational acceleration of more than \(10^4\times\) in this case. This demonstrates the effectiveness of the variational matrix-learning strategy for rapidly approximating oscillatory Helmholtz fields.

To further assess the reconstruction quality, cross-sectional visual comparisons of the predicted physical fields and their absolute error distributions are provided at the slice \(z=2\). Fig.~\ref{fig:helmholtz_poly_visual} shows the results for the Polynomial MMS, while Fig.~\ref{fig:helmholtz_cos_visual} presents the results for the high-frequency Cos-Cos-Sin MMS. In both cases, VMLFN accurately reproduces the main wave structures and yields localized error distributions, confirming its capability to represent oscillatory solutions with high computational efficiency.

\begin{table}
\centering
\caption{Quantitative Performance Comparison on 3D Heat Conduction (Adiabatic Top Surface). Evaluated against COMSOL Multiphysics simulations.}
\label{tab:heat_adiabatic_results}
\resizebox{\linewidth}{!}{\begin{tabular}{lccccc}
\toprule
\textbf{Source Type} & \textbf{Method} & \textbf{Avg Diff (K)} & \textbf{Max Diff (K)} & \textbf{Time (s)} & \textbf{Speedup} \\ \midrule
\multirow{2}{*}{Gaussian} & COMSOL  & $-$ & $-$ & $6$ & $1\times$ \\
& PINN  & $0.1510$ & $1.0086$ & $472.18$ & $0.012\times$ \\
& VMLFN  & $\mathbf{0.3574}$ & $\mathbf{1.3308}$ & $\mathbf{0.06}$ & $\mathbf{100\times}$ \\ \midrule
\multirow{2}{*}{Actual Chip} & COMSOL & $-$ & $-$ & $6$ & $1\times$ \\
& PINN   & $0.3537$& $3.8404$ & $545.27$ & $0.011\times$ \\
& VMLFN  & $\mathbf{0.4575}$ & $\mathbf{3.7307}$ & $\mathbf{0.
69}$ & $\mathbf{8.7\times}$ \\
\bottomrule
\multicolumn{6}{l}{ $\omega_{\mathrm{max}}^* = 5.5$, $\beta = 10^{-2}$ (Gaussian), $\omega_{\mathrm{max}}^* = 5.50$, $\beta = 10^{-2}$ (Actual Chip).} \\
\end{tabular}}
\end{table}



\subsection{Case II: 3-D Steady-State Heat Conduction With an Adiabatic Top Surface}
\label{subsec:case_heat_adiabatic}

\begin{figure*}[!htb]
    \centering
    \subfigure[]{\includegraphics[width=0.2\linewidth]{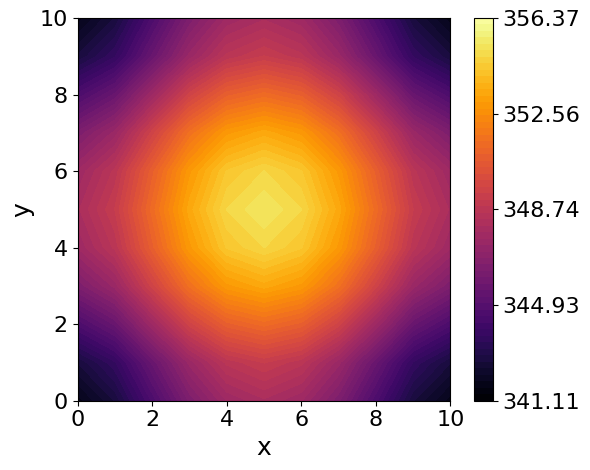}}%
    \subfigure[]{\includegraphics[width=0.2\linewidth]{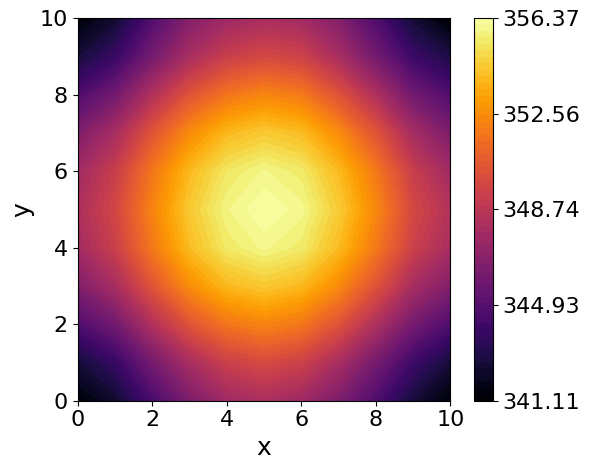}}
    \subfigure[]{\includegraphics[width=0.2\linewidth]{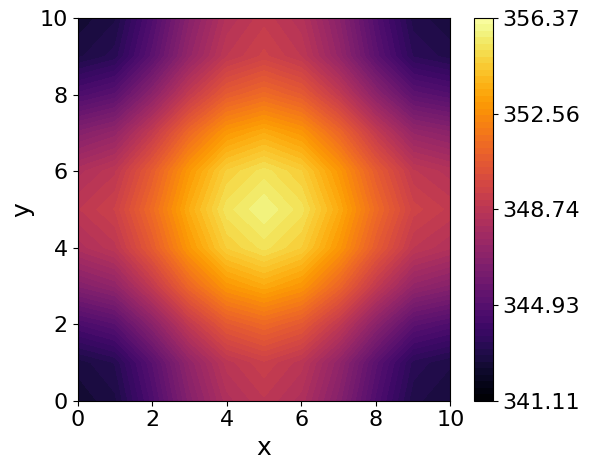}}%
    \subfigure[]{\includegraphics[width=0.2\linewidth]{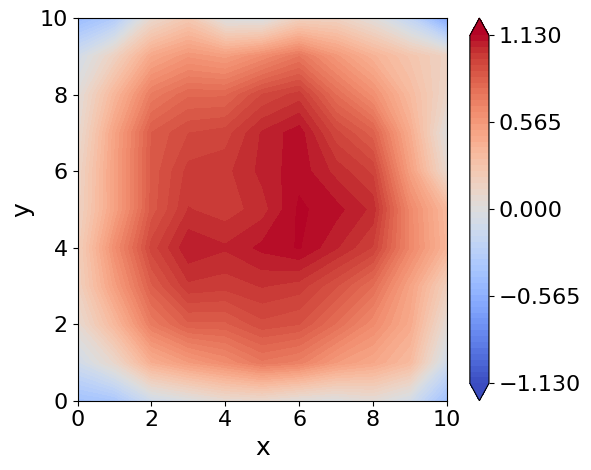}}%
    \subfigure[]{\includegraphics[width=0.2\linewidth]{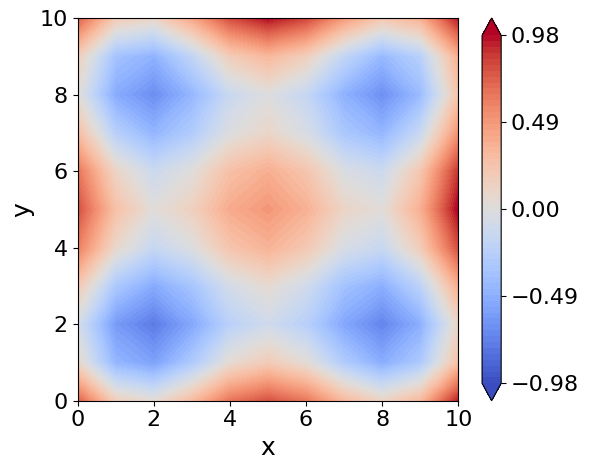}}\\%
    \subfigure[]{\includegraphics[width=0.2\linewidth]{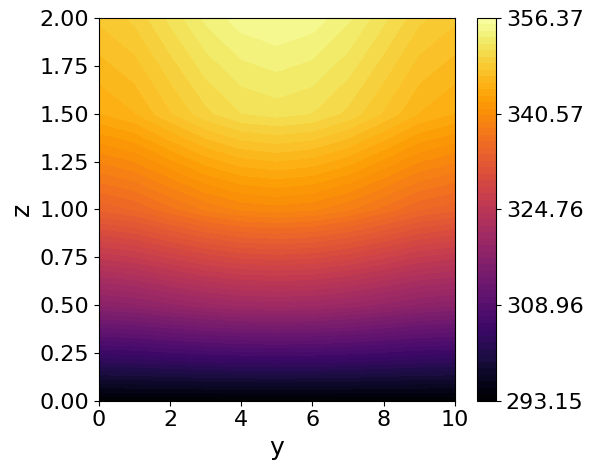}}%
    \subfigure[]{\includegraphics[width=0.2\linewidth]{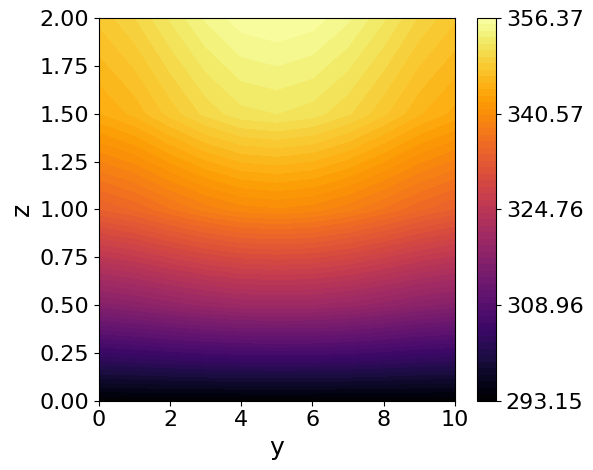}}%
    \subfigure[]{\includegraphics[width=0.2\linewidth]{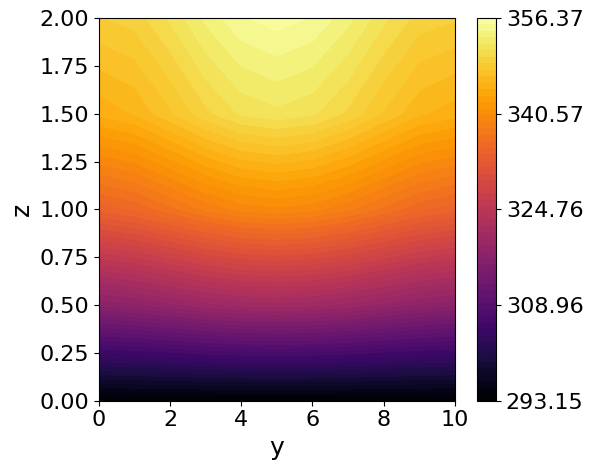}}%
    \subfigure[]{\includegraphics[width=0.2\linewidth]{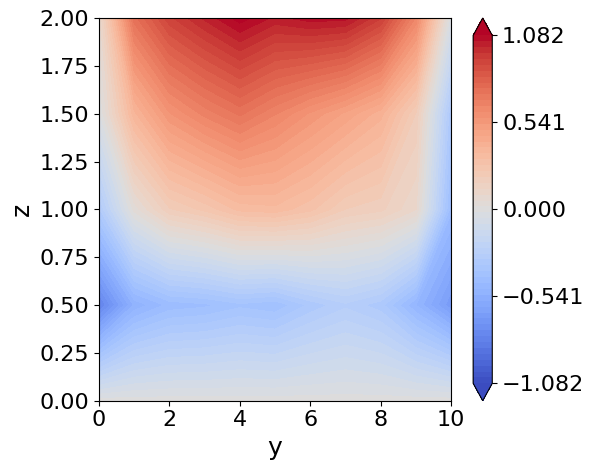}}%
    \subfigure[]{\includegraphics[width=0.2\linewidth]{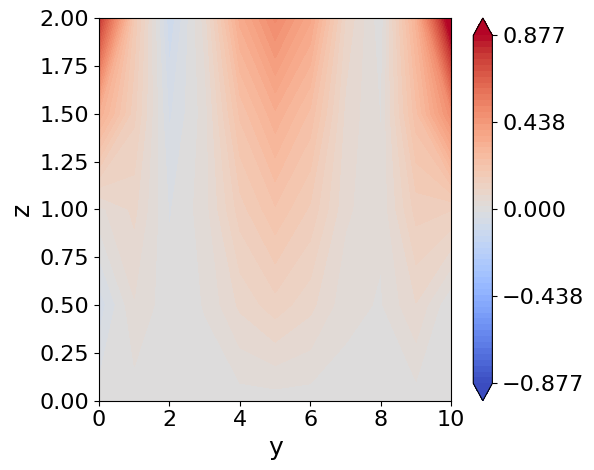}}%
    \caption{Temperature-field distributions for the Gaussian heat-source problem on two representative cross-sectional slices. 
(a)--(e) Results on the slice \(z=2\): (a) COMSOL reference solution, (b) VMLFN prediction, (c) PINN prediction, (d) absolute error of VMLFN, and (e) absolute error of PINN. 
(f)--(j) Results on the slice \(x=5\): (f) COMSOL reference solution, (g) VMLFN prediction, (h) PINN prediction, (i) absolute error of VMLFN, and (j) absolute error of PINN. }
    \label{fig:heat_gauss_visual}
\end{figure*}

    
\begin{figure*}[!htb]
    \centering
    \subfigure[]{\includegraphics[width=0.2\textwidth]{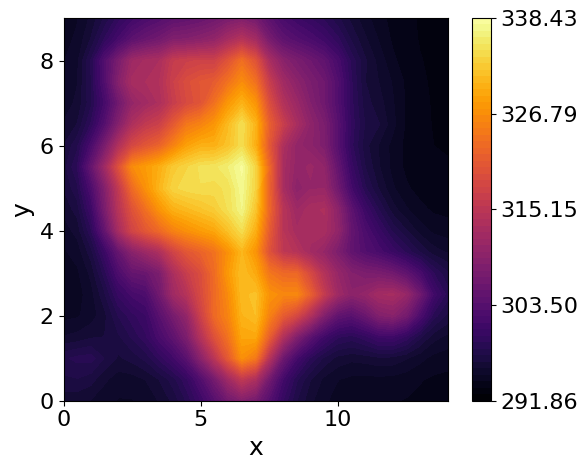}}%
    \subfigure[]{\includegraphics[width=0.2\textwidth]{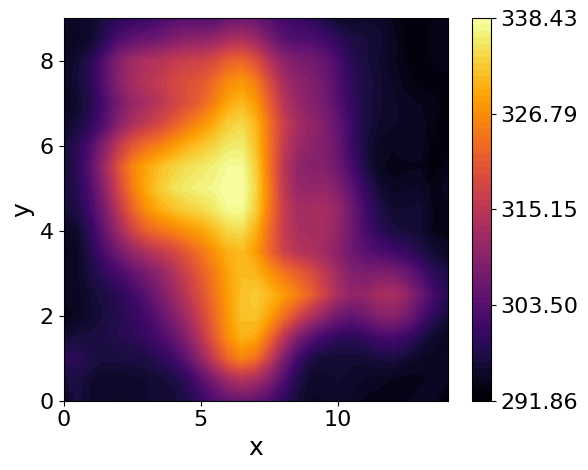}}%
    \subfigure[]{\includegraphics[width=0.2\textwidth]{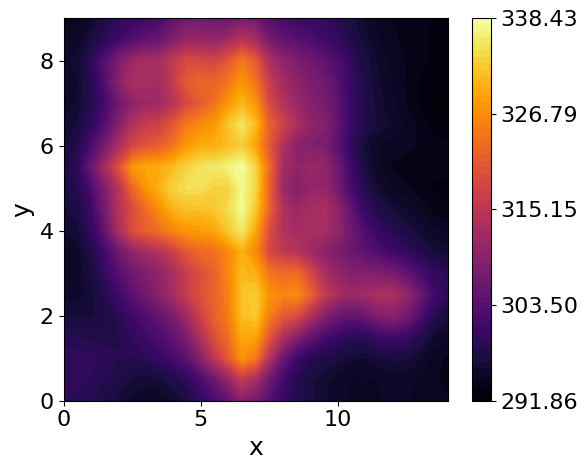}}%
    \subfigure[]{\includegraphics[width=0.2\textwidth]{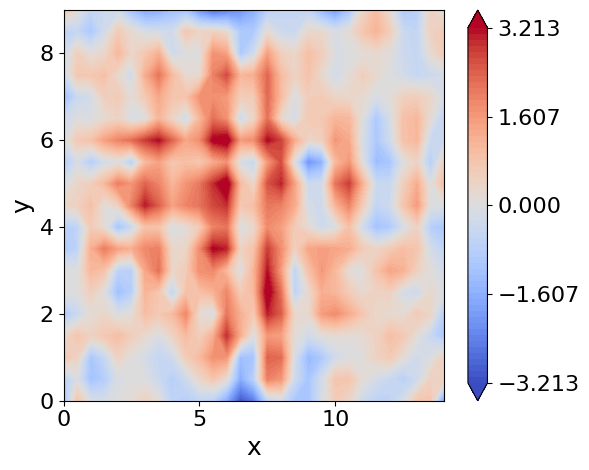}}%
    \subfigure[]{\includegraphics[width=0.2\textwidth]{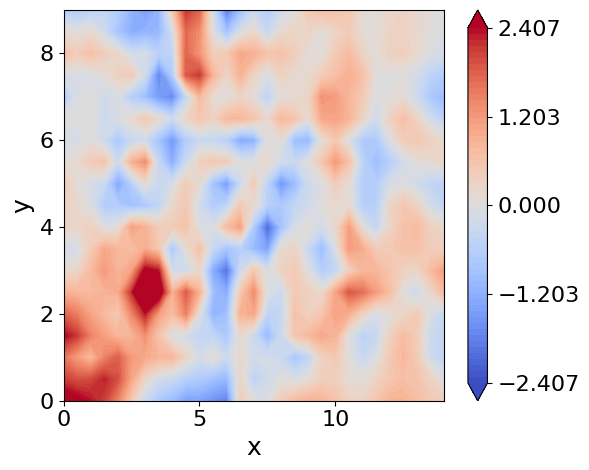}}\\%
    \subfigure[]{\includegraphics[width=0.2\textwidth]{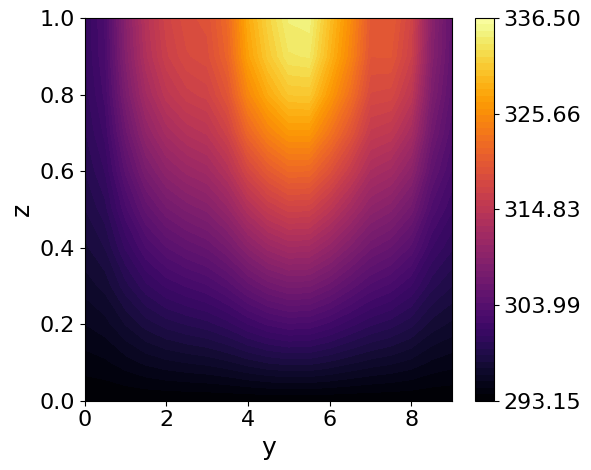}}%
    \subfigure[]{\includegraphics[width=0.2\textwidth]{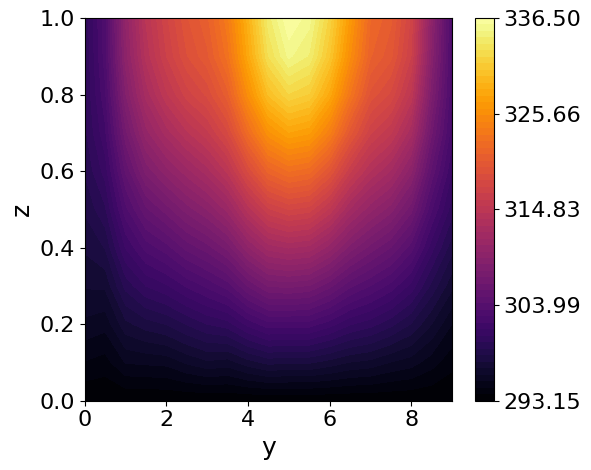}}%
    \subfigure[]{\includegraphics[width=0.2\textwidth]{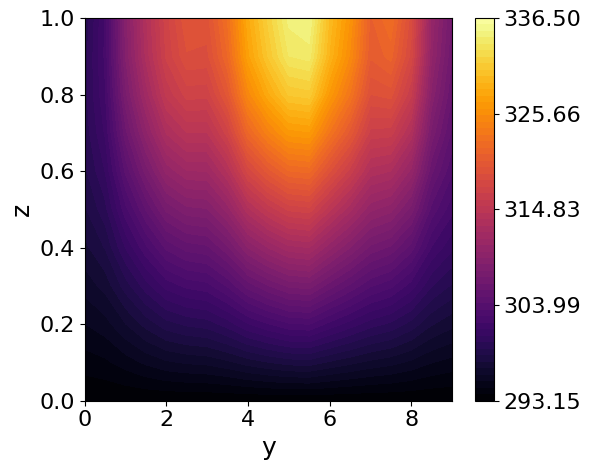}}%
    \subfigure[]{\includegraphics[width=0.2\textwidth]{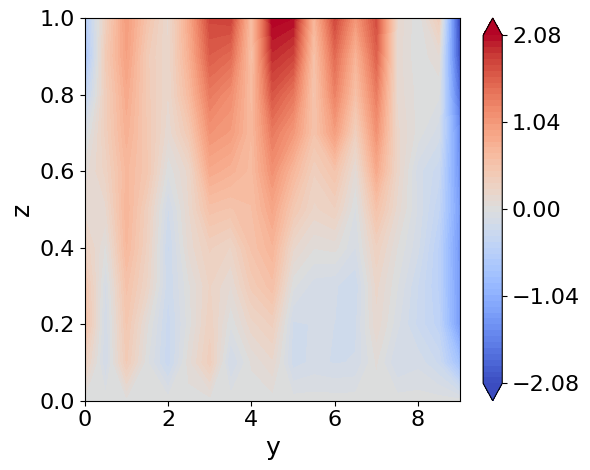}}%
    \subfigure[]{\includegraphics[width=0.2\textwidth]{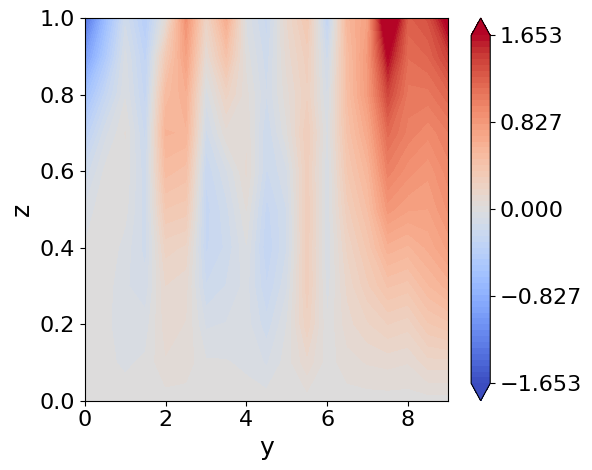}}%
    \caption{Temperature-field distributions for the actual chip heat-source problem on two representative cross-sectional slices. 
(a)--(e) Results on the slice \(z=1\): (a) COMSOL reference solution, (b) VMLFN prediction, (c) PINN prediction, (d) absolute error of VMLFN, and (e) absolute error of PINN. 
(f)--(j) Results on the slice \(x=5\): (f) COMSOL reference solution, (g) VMLFN prediction, (h) PINN prediction, (i) absolute error of VMLFN, and (j) absolute error of PINN.}
    \label{fig:heat_chip_visual}
\end{figure*}

We next consider the 3-D steady-state heat conduction problem. In contrast to the manufactured Helmholtz examples, practical thermal-management problems generally do not admit analytical solutions, especially when realistic and spatially complex heat-source layouts are considered. Therefore, the predictions of VMLFN and the conventional PINN are evaluated against high-fidelity finite-element reference solutions obtained from COMSOL Multiphysics.

In this case, the bottom surface is prescribed with a constant temperature $T_{\rm ref}=293.15~{\rm K}$ corresponding to a Dirichlet boundary condition. All remaining surfaces, including the top surface, are assumed to be perfectly insulated, i.e., adiabatic Neumann boundary conditions are imposed. Two representative heat-source configurations are investigated to assess the accuracy and robustness of the solvers.

The first configuration is a \textit{Gaussian heat source}, which produces a localized thermal hotspot and sharp temperature gradients. The thermal conductivity is set to $\kappa=60.0~{\rm W/(m\cdot K)} $. The second configuration is an \textit{actual chip heat source}, consisting of multiple localized heat-generation regions that mimic a realistic electronic-packaging layout. For this case, the thermal conductivity is set to $\kappa=131.0~{\rm W/(m\cdot K)}$.

\subsubsection{Model Configuration}

For the Gaussian heat-source case, VMLFN employs an ultra-compact representation with only 64 Fourier basis functions. Although the number of basis functions is small, the variational formulation allows the dominant thermal response to be efficiently captured through matrix projection. For the more complex actual chip heat-source case, the hidden dimension is increased to $ N_h=8000$
in order to resolve the stronger spatial heterogeneity induced by the multi-source layout. During the automatic tuning stage, the upper bound of the Fourier bandwidth, \(\omega_{\max}\), is selected by scanning candidate values with a reduced collocation-point budget, thereby limiting the additional cost of hyperparameter selection.

The conventional PINN baseline uses the same MLP architecture as in Case I, consisting of four hidden layers with 128 neurons per layer and the \(\tanh\) activation function. This architecture contains 50,177 trainable parameters. The network is trained using the Adam optimizer for a large number of epochs; for the actual chip heat-source scenario, up to 15,000 epochs are used. The total loss consists of the strong-form PDE residual and boundary-condition penalties, with empirically tuned weights for the Dirichlet and Neumann terms.

\subsubsection{Quantitative and Visual Comparison}

As reported in Table~\ref{tab:heat_adiabatic_results}, the PINN baseline considered here is a boundary-hard-constrained PINN rather than a conventional soft-constrained PINN. Specifically, the essential Dirichlet boundary condition is embedded analytically into the trial solution, which avoids explicit penalty enforcement of this boundary condition and partly explains the favorable accuracy of the PINN results. By contrast, a traditional PINN typically needs to balance the PDE residual and boundary-condition losses during training, making the optimization process more sensitive to loss weighting and more difficult to stabilize for steady-state heat conduction problems.

For the Gaussian heat-source case, the hard-constrained PINN achieves an average temperature difference of 0.1510 K and a maximum difference of 1.0086 K, compared with 0.3574 K and 1.3308 K for VMLFN. For the actual-chip case, the PINN again yields a lower average difference (0.3537 K versus 0.4575 K), while VMLFN produces a slightly smaller maximum difference (3.7307 K versus 3.8404 K). However, these accuracy advantages of the PINN come at a much higher computational cost: it requires 472.18 s and 545.27 s for the two cases, whereas VMLFN completes the full computation in only 0.06 s and 0.69 s, respectively.

From a methodological perspective, VMLFN reformulates the problem in weak variational form and employs integration by parts, thereby avoiding direct minimization of the second-order strong-form residual. The Dirichlet condition on the bottom surface is imposed analytically through a distance-type envelope function, while the homogeneous Neumann conditions on the remaining surfaces are naturally incorporated into the variational formulation. Consequently, although the hard-constrained PINN attains slightly better accuracy in most metrics, VMLFN still delivers competitive predictive performance with dramatically reduced runtime.

Overall, compared with the PINN baseline, VMLFN is approximately \(7.9\times10^3\) times faster for the Gaussian case and about \(7.9\times10^2\) times faster for the actual-chip case. It also outperforms the COMSOL runtime, achieving a speedup of \(100\times\) for the Gaussian case and \(8.7\times\) for the actual-chip case. These results indicate that the proposed matrix-learning framework provides a highly efficient alternative for elliptic thermal problems, especially when rapid solution is of primary importance.

To further illustrate the prediction quality, cross-sectional temperature distributions and absolute error maps are presented in Figs.~\ref{fig:heat_gauss_visual} and~\ref{fig:heat_chip_visual}. Fig.~\ref{fig:heat_gauss_visual} shows the Gaussian heat-source results on representative slices, while Fig.~\ref{fig:heat_chip_visual} presents the corresponding results for the actual chip heat-source layout. The visual comparisons confirm that VMLFN closely reproduces the COMSOL reference fields and yields lower error concentrations than the conventional PINN, especially near localized heat-source regions.




\begin{table}
\centering
\caption{Quantitative Performance Comparison on 3D Heat Conduction with Top Heat Convection/Flux ($q_{top} = 10^6$ W/m$^2$). Evaluated against COMSOL Multiphysics.}
\label{tab:heat_flux_results}
\begin{tabular}{lccccc}
\toprule
\textbf{Method} & \textbf{Avg Diff (K)} & \textbf{Max Diff (K)} & \textbf{Time (s)} & \textbf{Speedup} \\ \midrule
 COMSOLl & $-$ & $-$  & $6$ & $1\times$ \\
PINN  &  $0.3129$ & $4.1760$ & $415.49$ & $0.014\times$ \\
VMLFN  &  $\mathbf{0.6300}$ & $\mathbf{3.1913}$ & $\mathbf{0.53}$ & $\mathbf{11.32\times}$ \\ \bottomrule
\multicolumn{5}{l}{\footnotesize $\omega_{\mathrm{max}}^* = 0.3535$, $\beta = 10^{-4}$.} \\
\end{tabular}
\end{table}



\subsection{Case III: Robustness to Nonhomogeneous Neumann Boundaries With Top-Surface Heat Flux}
\label{subsec:case_heat_flux}

In practical thermal-management applications, exposed surfaces are rarely perfectly adiabatic. Instead, they may be subjected to prescribed heat fluxes, convective cooling, or other nonhomogeneous boundary interactions. To examine the robustness of the proposed method under such conditions, we consider a heat-conduction problem in which a high constant heat flux is imposed on the top surface $q_{\rm top}=10^6~{\rm W/m^2}$.
The corresponding nonhomogeneous Neumann boundary condition is written as $\kappa \nabla T\cdot\mathbf{n}=q_{\rm top}$.

\begin{figure*}[!htb]
    \centering
    \subfigure[]{\includegraphics[width=0.2\textwidth]{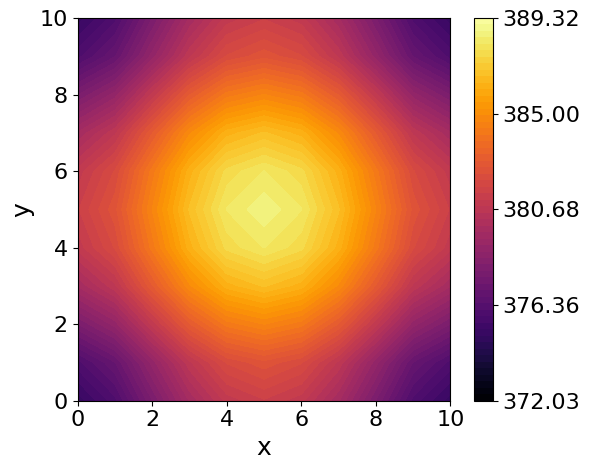}}%
    \subfigure[]{\includegraphics[width=0.2\textwidth]{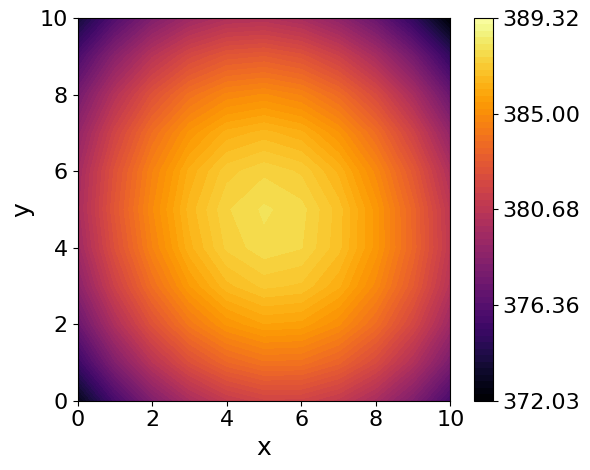}}%
    \subfigure[]{\includegraphics[width=0.2\textwidth]{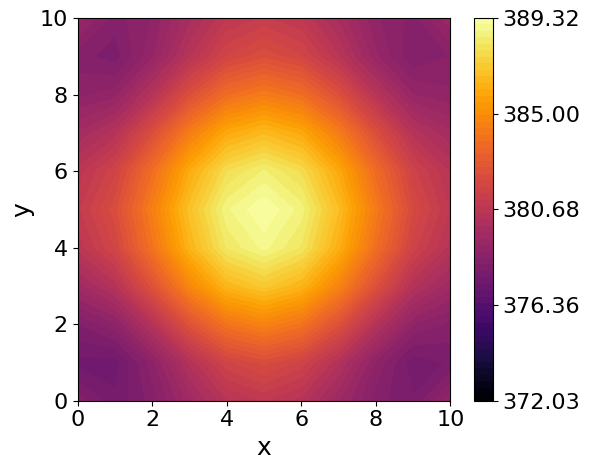}}%
    \subfigure[]{\includegraphics[width=0.2\textwidth]{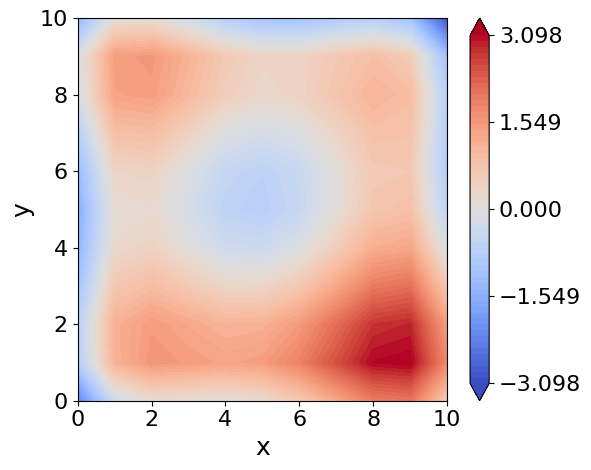}}%
    \subfigure[]{\includegraphics[width=0.2\textwidth]{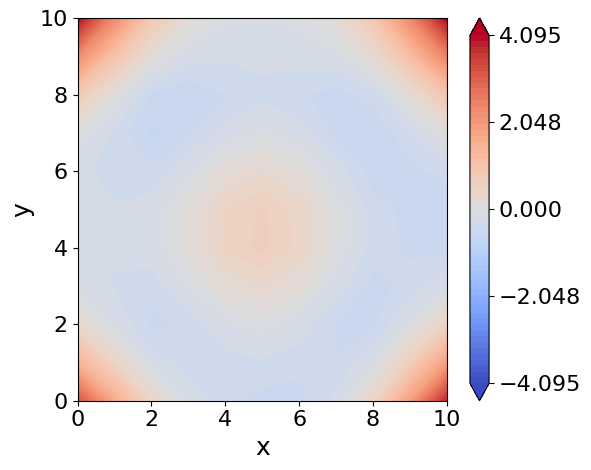}}%
    \caption{Temperature-field distributions for the heat-source problem on the representative slice \(z=2\). 
(a) COMSOL reference solution, (b) VMLFN prediction, (c) PINN prediction, (d) absolute error of VMLFN, and (e) absolute error of PINN.}
    \label{fig:heat_flux_zvisual}
\end{figure*}
\begin{figure*}[!htb]
    \centering
    \subfigure[]{\includegraphics[width=0.2\textwidth]{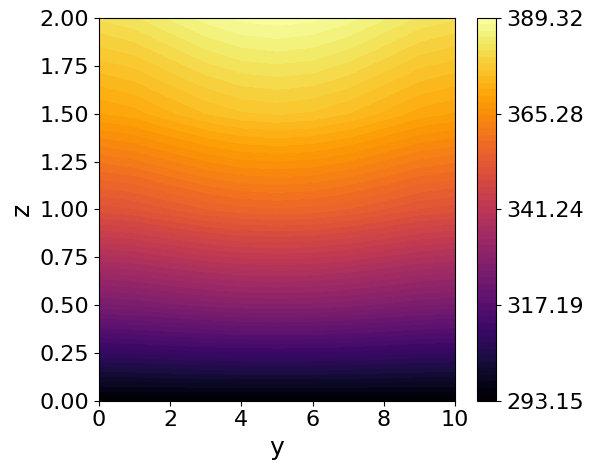}}%
    \subfigure[]{\includegraphics[width=0.2\textwidth]{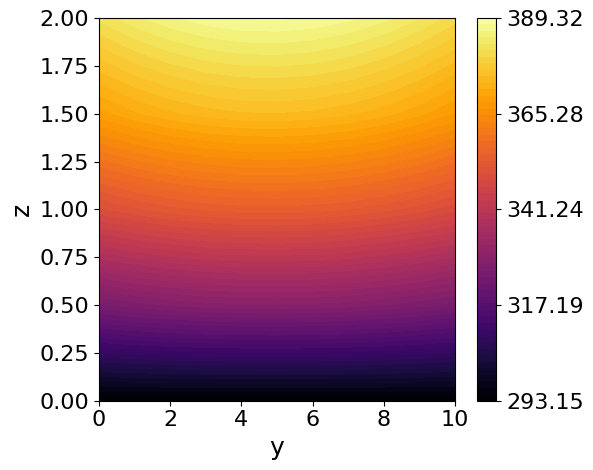}}%
    \subfigure[]{\includegraphics[width=0.2\textwidth]{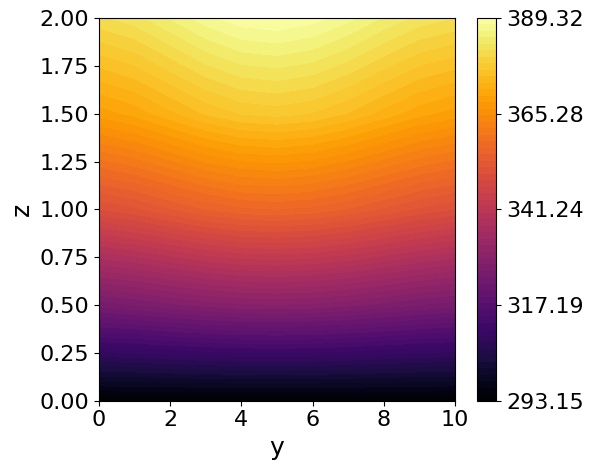}}%
    \subfigure[]{\includegraphics[width=0.2\textwidth]{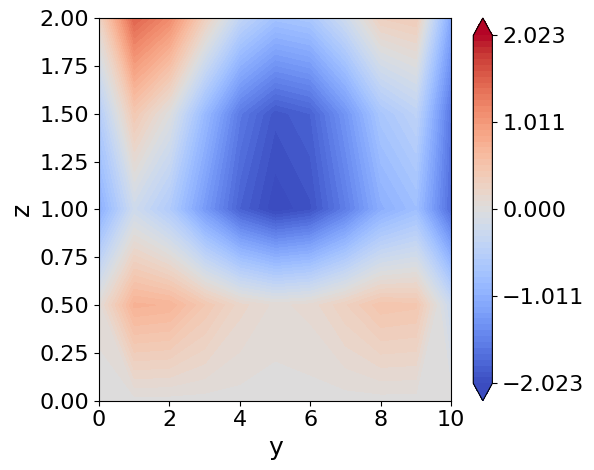}}%
    \subfigure[]{\includegraphics[width=0.2\textwidth]{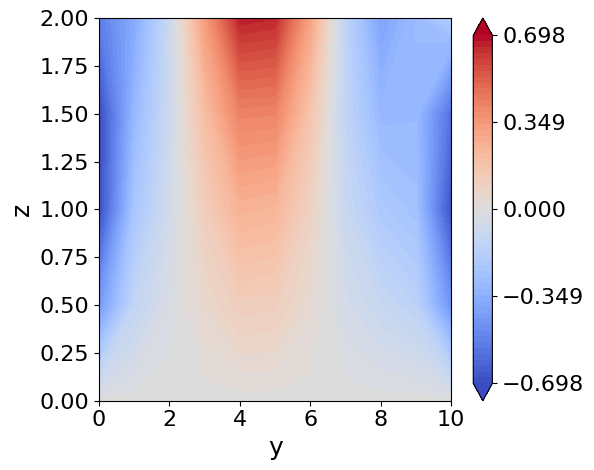}}%
    \caption{Temperature-field distributions for the heat-source problem on the representative slice \(x=5\). 
(a) COMSOL reference solution, (b) VMLFN prediction, (c) PINN prediction, (d) absolute error of VMLFN, and (e) absolute error of PINN.}
    \label{fig:heat_flux_xvisual}
\end{figure*}

For conventional PINNs, such nonhomogeneous Neumann conditions are often difficult to enforce accurately. Since the boundary flux involves first-order spatial derivatives of the neural-network output, its penalty-based enforcement through automatic differentiation may introduce strong competition between the PDE residual loss, the Dirichlet boundary loss, and the Neumann boundary loss. This imbalance can lead to unstable optimization and localized boundary-gradient errors.

In contrast, VMLFN incorporates the prescribed top-surface heat flux directly into the global load vector \(\mathbf{F}\) through the boundary integral arising from the weak formulation. Therefore, the nonhomogeneous Neumann condition is treated as a natural boundary condition, without requiring explicit gradient matching or manually tuned Neumann penalty weights. This property is particularly advantageous for heat-transfer problems involving strong boundary fluxes.

Table~\ref{tab:heat_flux_results} summarizes the quantitative performance for the top-surface heat-flux case. The conventional PINN achieves a lower average temperature difference of \(0.3129\) K; however, it requires \(415.49\) s of training and produces a relatively large maximum error of \(4.1760\) K. Its computational speed is only \(0.014\times\) that of the COMSOL baseline. By comparison, VMLFN completes the entire computation in \(0.53\) s, corresponding to an \(11.32\times\) speedup over COMSOL. Although its average temperature difference is \(0.6300\) K, its maximum temperature difference is reduced to \(3.1913\) K, indicating improved control of the worst-case error.

These results show that VMLFN offers a substantially more efficient solution strategy for heat-conduction problems with nonhomogeneous Neumann boundaries. By embedding the prescribed flux into the variational load vector, the method avoids the loss-balancing difficulty encountered by PINNs and maintains competitive prediction accuracy while significantly reducing runtime.

The corresponding cross-sectional temperature profiles and absolute error distributions are visualized in Figs.~\ref{fig:heat_flux_zvisual} and~\ref{fig:heat_flux_xvisual}. Fig.~\ref{fig:heat_flux_zvisual} presents the results on the slice \(z=2\), while Fig.~\ref{fig:heat_flux_xvisual} shows the corresponding results on the slice \(x=5\).

\begin{figure*}[!hb]
    \centering
    \subfigure[]{\includegraphics[width=0.16\textwidth]{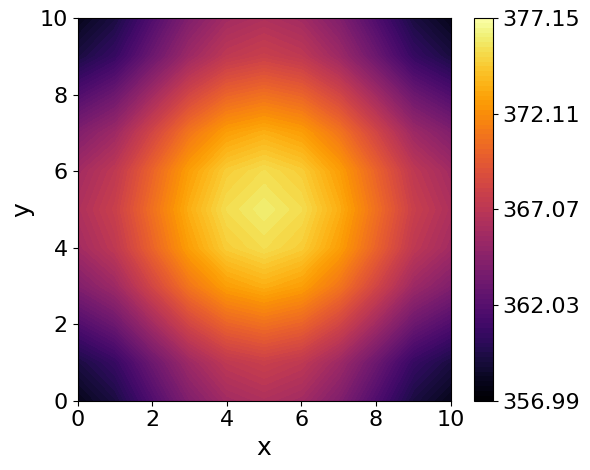}}%
    \subfigure[]{\includegraphics[width=0.16\textwidth]{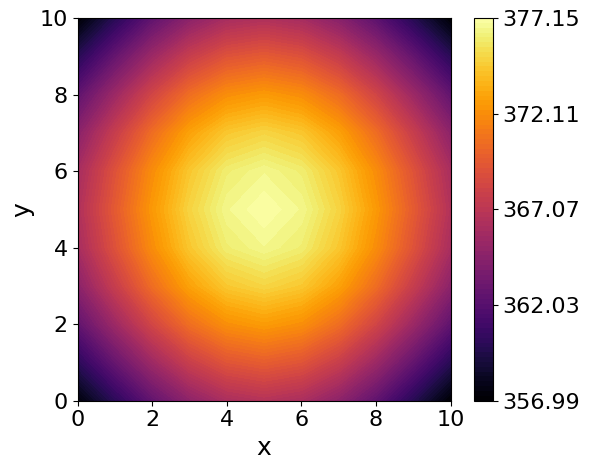}}%
    \subfigure[]{\includegraphics[width=0.16\textwidth]{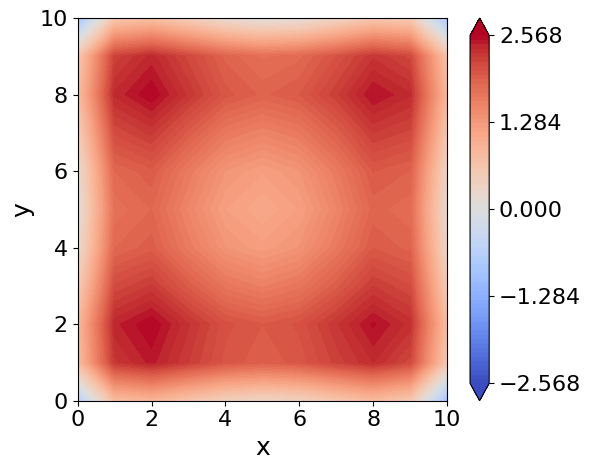}}%
    \subfigure[]{\includegraphics[width=0.16\textwidth]{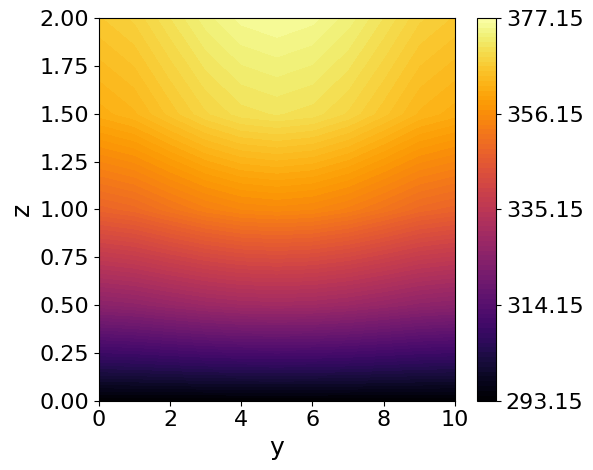}}%
    \subfigure[]{\includegraphics[width=0.16\textwidth]{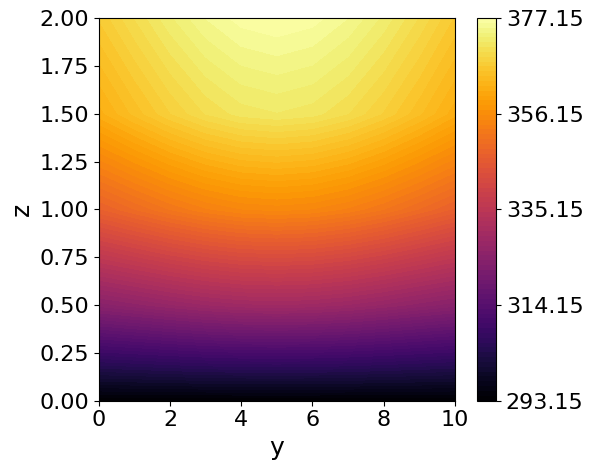}}%
    \subfigure[]{\includegraphics[width=0.16\textwidth]{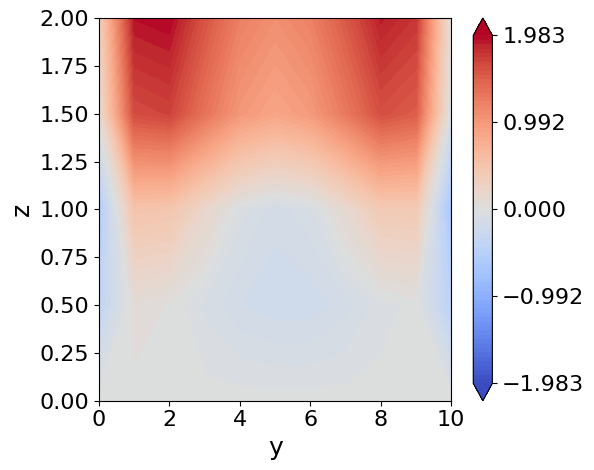}}%
    \caption{Temperature-field distributions for the heat-source problem with \(\kappa=45\) on two representative cross-sectional slices. 
(a)--(c) Results on the slice \(z=2\): (a) COMSOL reference solution, (b) VMLFN prediction, and (c) absolute error of VMLFN. 
(d)--(f) Results on the slice \(x=5\): (d) COMSOL reference solution, (e) VMLFN prediction, and (f) absolute error of VMLFN.}
    \label{fig:k45}
\end{figure*}
\begin{figure*}[!hb]
    \centering
    \subfigure[]{\includegraphics[width=0.16\textwidth]{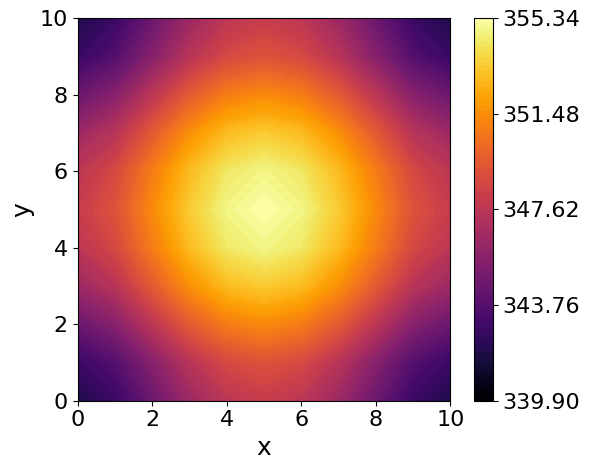}}%
    \subfigure[]{\includegraphics[width=0.16\textwidth]{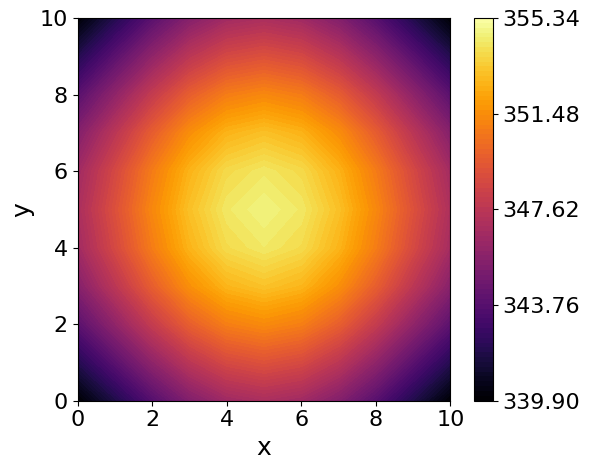}}%
    \subfigure[]{\includegraphics[width=0.16\textwidth]{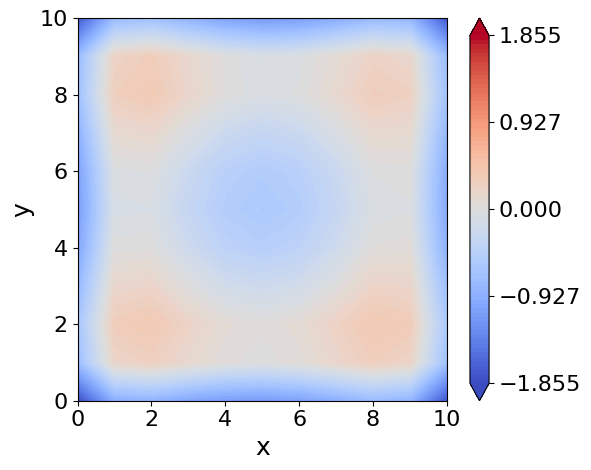}}%
    \subfigure[]{\includegraphics[width=0.16\textwidth]{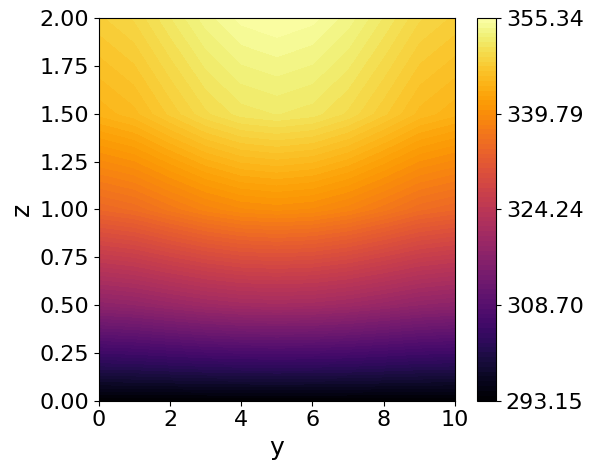}}%
    \subfigure[]{\includegraphics[width=0.16\textwidth]{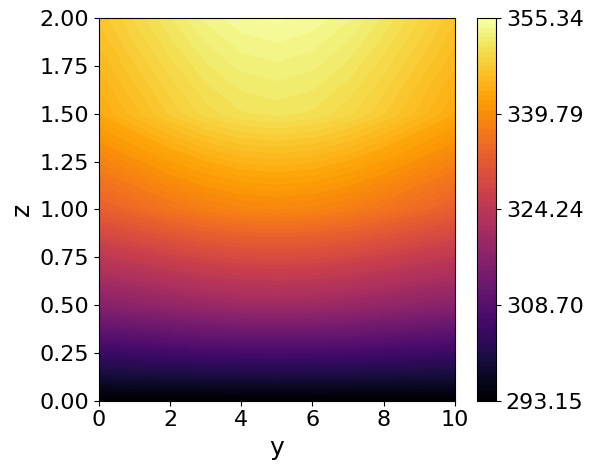}}%
    \subfigure[]{\includegraphics[width=0.16\textwidth] {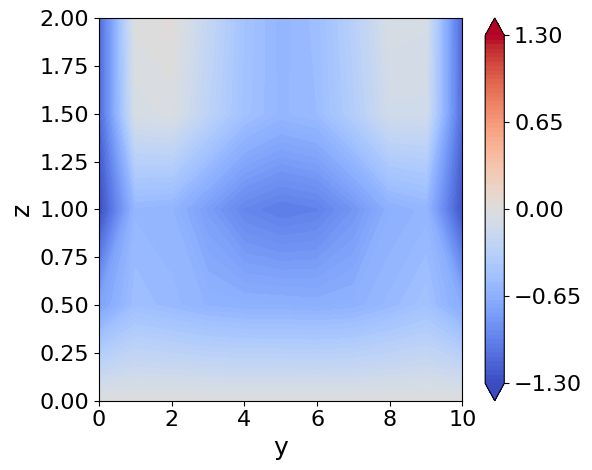}}%
    \caption{Temperature-field distributions for the heat-source problem with \(\kappa=60\) on two representative cross-sectional slices. 
(a)--(c) Results on the slice \(z=2\): (a) COMSOL reference solution, (b) VMLFN prediction, and (c) absolute error of VMLFN. 
(d)--(f) Results on the slice \(x=5\): (d) COMSOL reference solution, (e) VMLFN prediction, and (f) absolute error of VMLFN.}
    \label{fig:k60}
\end{figure*}
\begin{figure*}[!hb]
    \centering
    \subfigure[]{\includegraphics[width=0.16\textwidth]{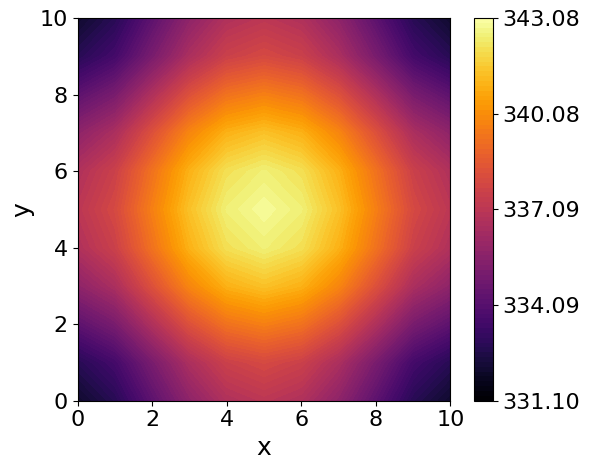}}%
    \subfigure[]{\includegraphics[width=0.16\textwidth]{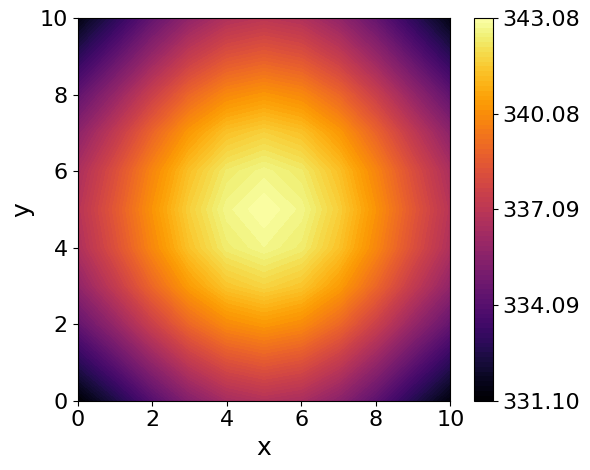}}%
    \subfigure[]{\includegraphics[width=0.16\textwidth]{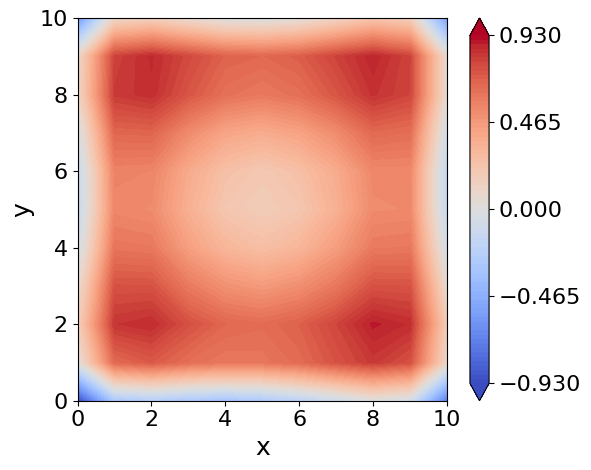}}%
    \subfigure[]{\includegraphics[width=0.16\textwidth]{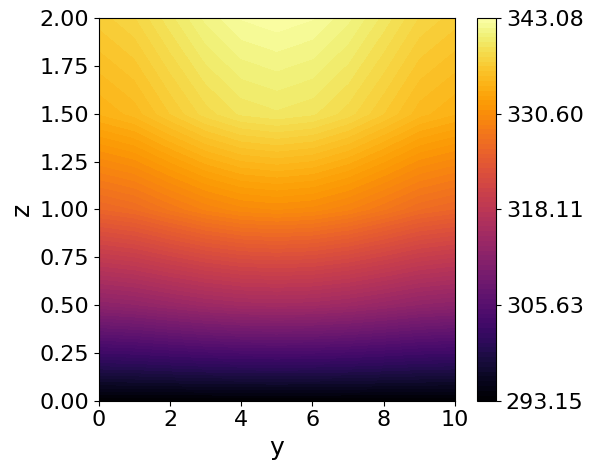}}%
    \subfigure[]{\includegraphics[width=0.16\textwidth]{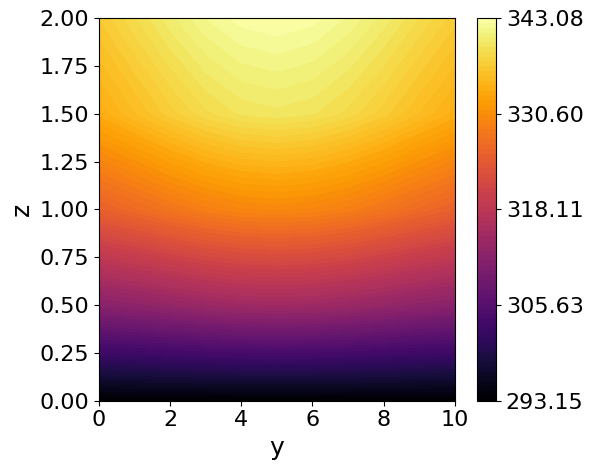}}%
    \subfigure[]{\includegraphics[width=0.16\textwidth]{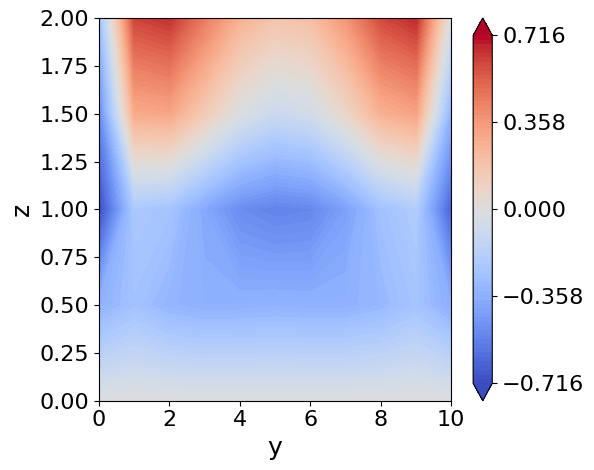}}%
    \caption{Temperature-field distributions for the heat-source problem with \(\kappa=75\) on two representative cross-sectional slices. 
(a)--(c) Results on the slice \(z=2\): (a) COMSOL reference solution, (b) VMLFN prediction, and (c) absolute error of VMLFN. 
(d)--(f) Results on the slice \(x=5\): (d) COMSOL reference solution, (e) VMLFN prediction, and (f) absolute error of VMLFN.}
    \label{fig:k75}
\end{figure*}





\begin{table}
\centering
\caption{Zero-Shot Inference Performance of the 4D Spatio-Parametric Model on Unseen Conductivities. (Inference evaluated on 605 3D spatial points per pass).}
\label{tab:parametric_results}
\resizebox{\linewidth}{!}{\begin{tabular}{ccccc}
\toprule
\textbf{$\kappa$} & \textbf{Avg Diff (K)} & \textbf{Max Diff (K)} & \textbf{Time(s)} & \textbf{Speed/Point} \\ \midrule
$45.0$ W/(m$\cdot$K) & $0.6752$ & $2.5682$ & $0.308$ ms & $0.51 \ \mu$s \\
$60.0$ W/(m$\cdot$K) & $0.4786$ & $1.8996$ & $0.319$ ms & $0.53 \ \mu$s \\
$75.0$ W/(m$\cdot$K) & $0.3197$ & $1.1660$ & $0.317$ ms & $0.52 \ \mu$s \\ \bottomrule
\end{tabular}}
\end{table}


\subsection{Case IV: 4-D Spatio-Parametric Surrogate Modeling for Material Design}
\label{subsec:case_parametric_surrogate}

In practical engineering design, material optimization and parametric analysis often require repeated solutions of the governing PDEs over a broad range of physical properties. For conventional numerical solvers, such as FEM, each new material parameter generally requires a separate system assembly and solution. Similarly, standard PINNs typically require retraining or at least substantial fine-tuning when the physical parameter changes. These repeated computations can become prohibitively expensive in large-scale design exploration.

To address this limitation, the proposed VMLFN framework is extended to a 4-D spatio-parametric surrogate model by treating the thermal conductivity \(\kappa\) as an additional continuous input coordinate. Specifically, the network approximates the temperature field as $T = T(x,y,z,\kappa)$, where the conductivity is defined over the design interval $\kappa \in [30.0,90.0]~{\rm W/(m\cdot K)} $.
By embedding \(\kappa\) directly into the input space, the model learns the global dependence of the thermal response on both spatial coordinates and material properties.

During training, the optimal Fourier frequency bandwidth is selected using the heuristic frequency-scanning strategy described in Algorithm~\ref{alg:HFS}. Once the bandwidth is determined, it remains fixed for the construction of the global surrogate. Remarkably, only 10 discrete conductivity snapshots are used to sample the parametric domain. With these sparse parameter samples, the proposed method assembles a single global variational system and performs only one matrix inversion to obtain the universal output-weight matrix. In this case, the hidden dimension is set to $N_h=2000$.
Thus, unlike conventional solvers that repeatedly solve independent problems for different conductivities, VMLFN constructs a unified surrogate over the entire conductivity interval.

To rigorously assess the zero-shot generalization capability of the learned spatio-parametric surrogate, the trained model is directly queried at three unseen conductivity values $
\kappa=45.0,\quad 60.0,\quad 75.0~{\rm W/(m\cdot K)}$.
These testing values are not included in the training snapshots. Therefore, the evaluation directly measures whether the model has learned a continuous and physically consistent mapping with respect to the material parameter.

\begin{figure*}[!hb]
    \centering
    \setlength{\tabcolsep}{1pt}
    \renewcommand{\arraystretch}{1.05}

    \begin{tabular}{C{0.08\textwidth} C{0.17\textwidth} C{0.17\textwidth} C{0.17\textwidth} C{0.17\textwidth} C{0.17\textwidth}}
        \small\textbf{Case 1} &
        \includegraphics[width=0.17\textwidth]{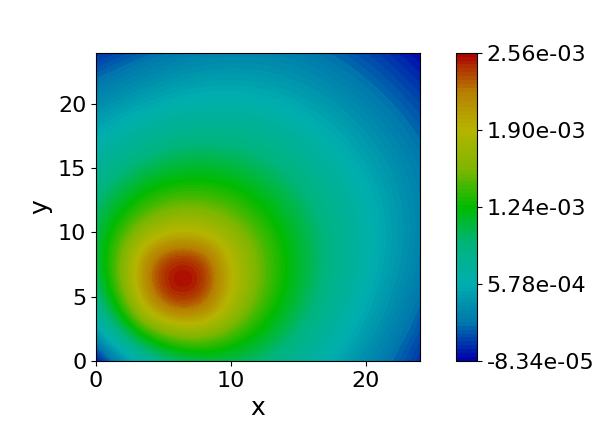} &
        \includegraphics[width=0.17\textwidth]{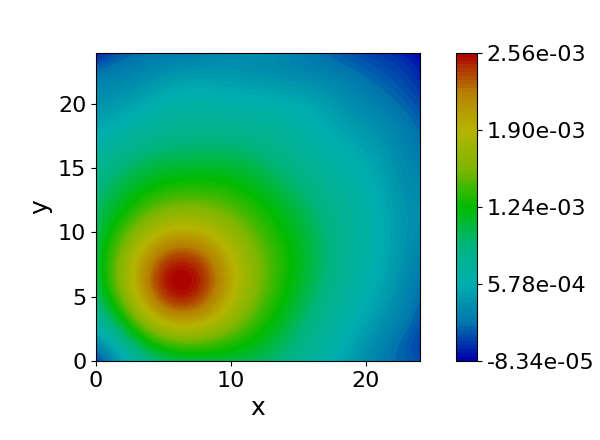} &
        \includegraphics[width=0.17\textwidth]{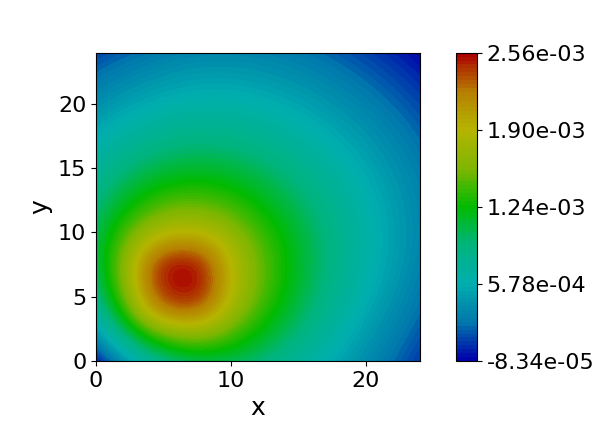} &
        \includegraphics[width=0.17\textwidth]{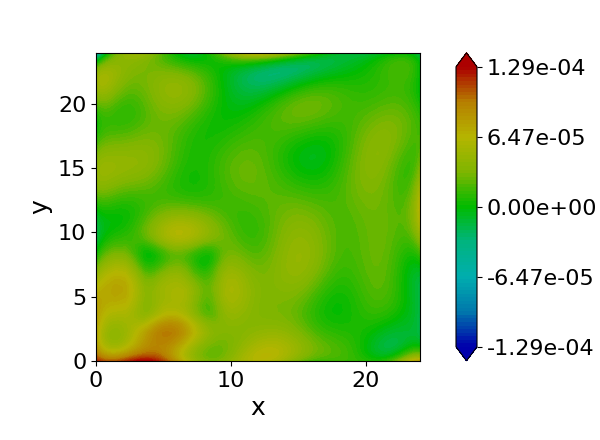} &
        \includegraphics[width=0.17\textwidth]{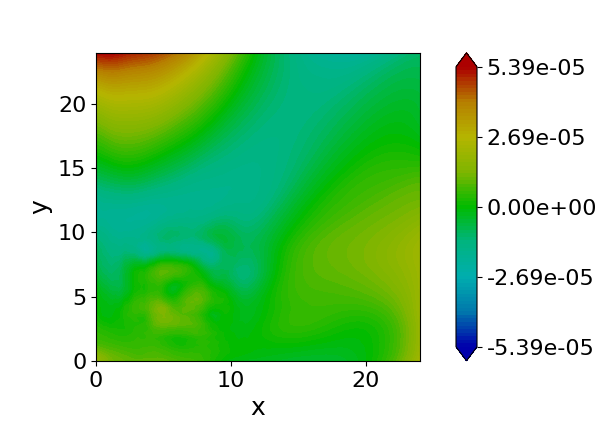} \\[1ex]

        \small\textbf{Case 2} &
        \includegraphics[width=0.17\textwidth]{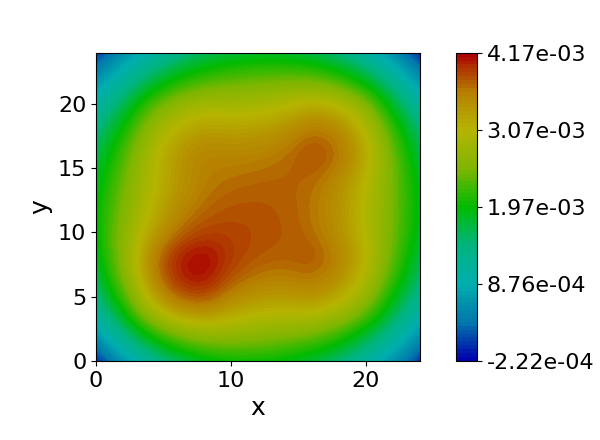} &
        \includegraphics[width=0.17\textwidth]{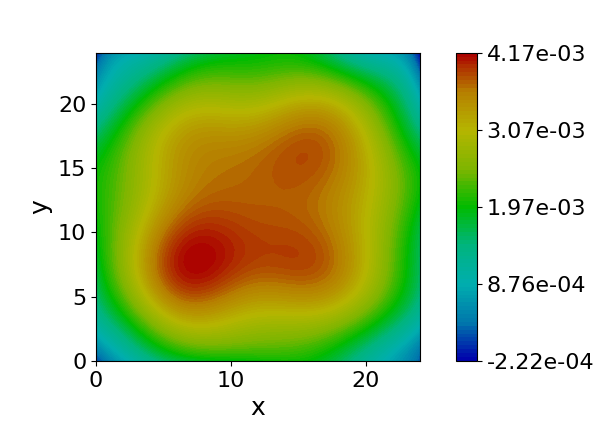} &
        \includegraphics[width=0.17\textwidth]{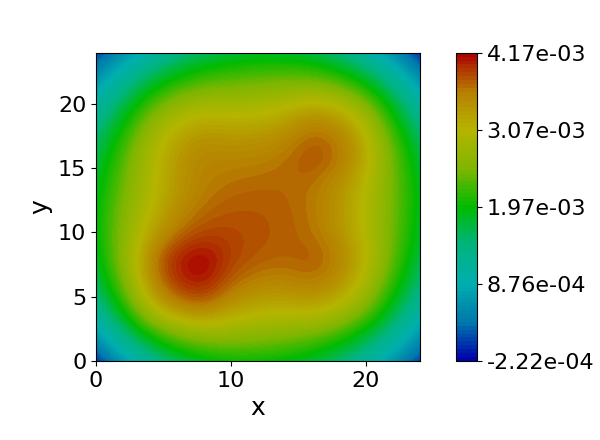} &
        \includegraphics[width=0.17\textwidth]{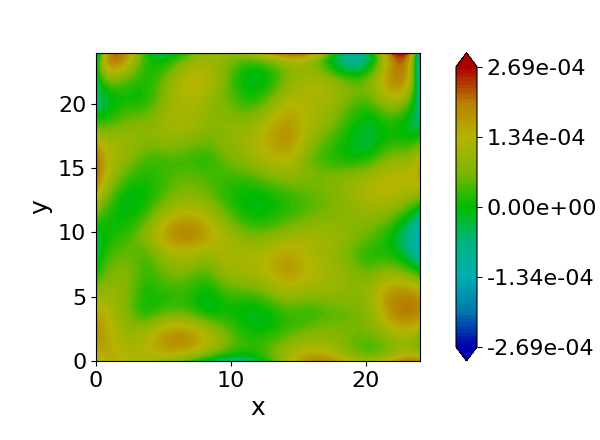} &
        \includegraphics[width=0.17\textwidth]{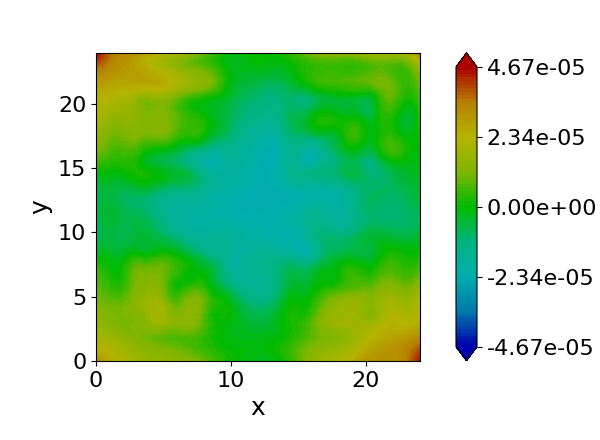} \\[1ex]

        \small\textbf{Case 3} &
        \includegraphics[width=0.17\textwidth]{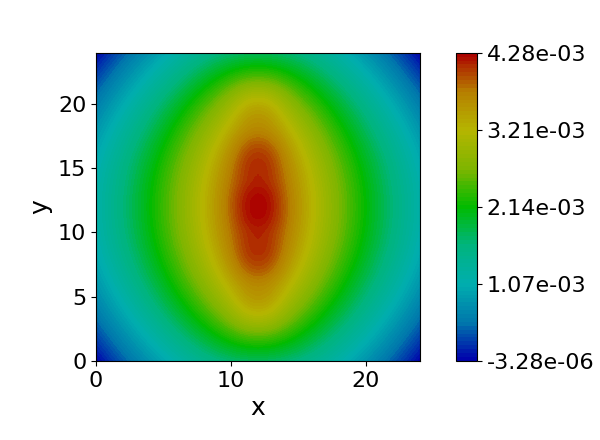} &
        \includegraphics[width=0.17\textwidth]{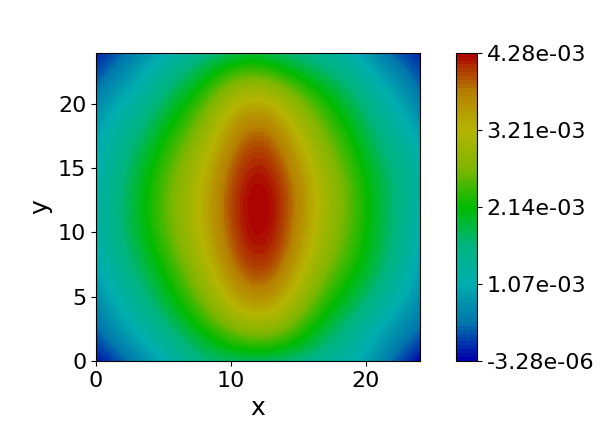} &
        \includegraphics[width=0.17\textwidth]{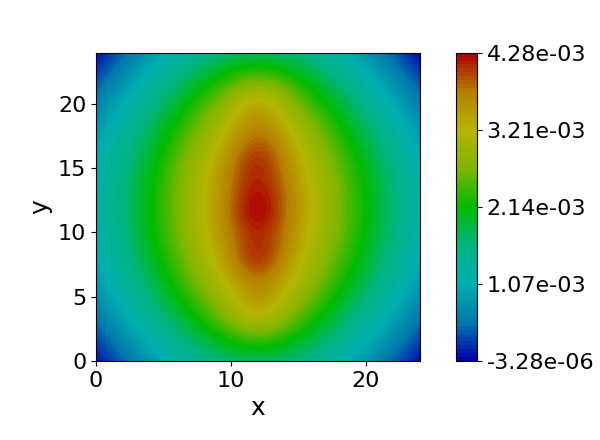} &
        \includegraphics[width=0.17\textwidth]{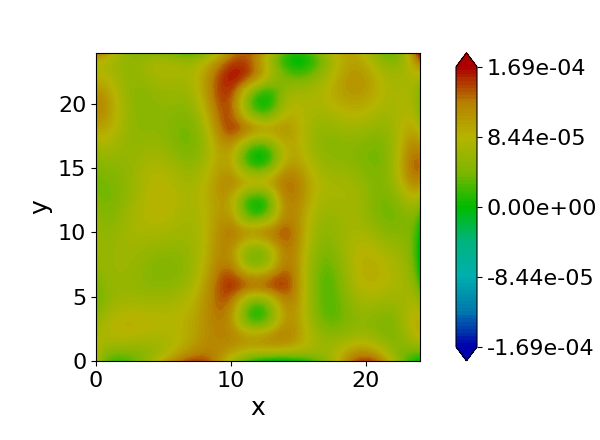} &
        \includegraphics[width=0.17\textwidth]{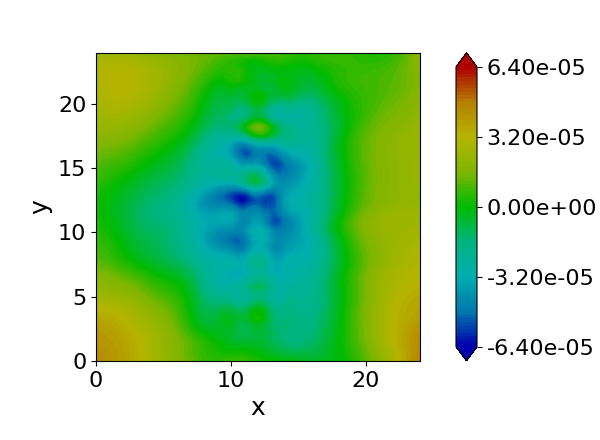} \\[1ex]

        \small\textbf{Case 4} &
        \includegraphics[width=0.17\textwidth]{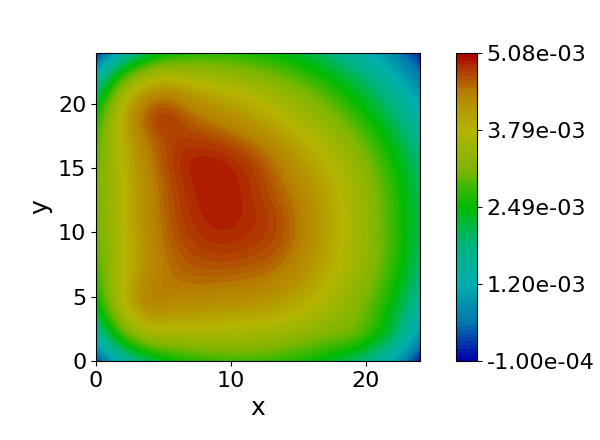} &
        \includegraphics[width=0.17\textwidth]{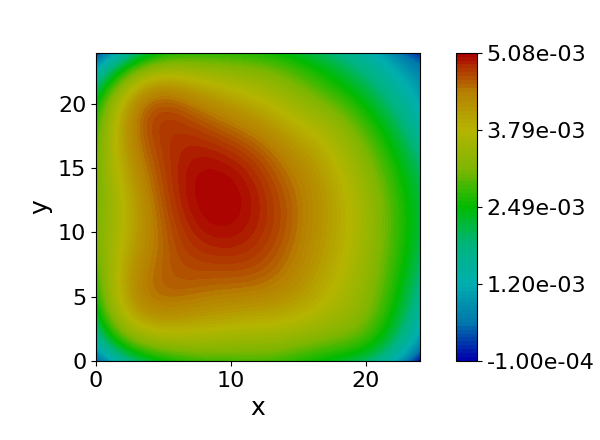} &
        \includegraphics[width=0.17\textwidth]{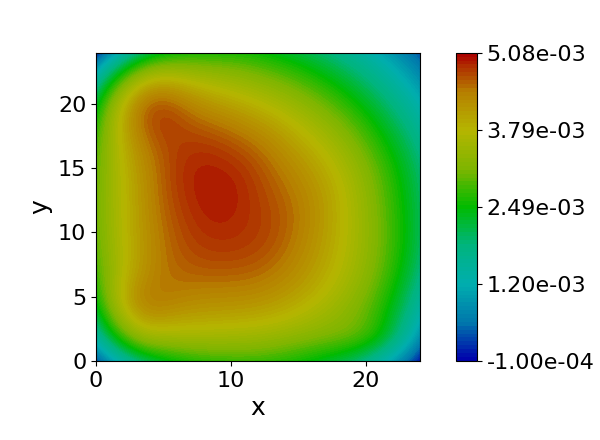} &
        \includegraphics[width=0.17\textwidth]{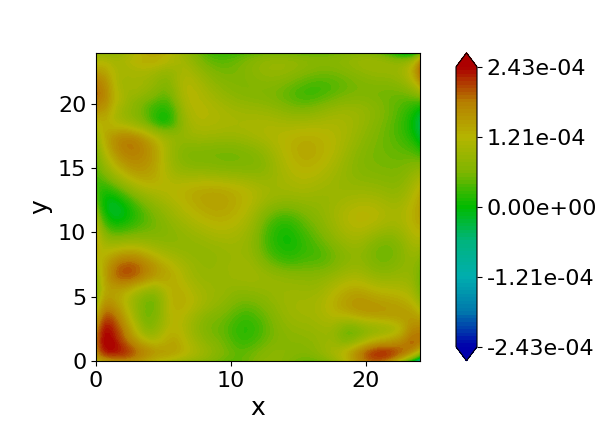} &
        \includegraphics[width=0.17\textwidth]{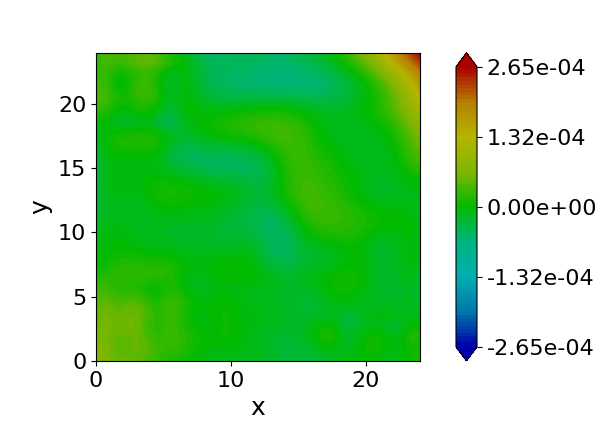} \\[0.8ex]

        &
        \small (a)&
        \small (b)&
        \small (c)&
        \small (d)&
        \small (e)\\
    \end{tabular}

    \caption{Warpage distributions of the 3-D thermoelastic deformation field in the heterogeneous composite, represented by the out-of-plane displacement component \(u_z\). 
(a) COMSOL reference solution, (b) VMLFN prediction, (c) Var. PINN prediction, (d) absolute error of VMLFN, and (e) absolute error of Var. PINN.}
    \label{fig:warping_results_all}
\end{figure*}

As reported in Table~\ref{tab:parametric_results}, the proposed surrogate model achieves accurate predictions at all three unseen conductivities, demonstrating strong interpolation capability across the material-design space. More importantly, once the universal weight matrix has been obtained, predicting a new thermal configuration no longer requires PDE assembly, matrix inversion, or iterative optimization. The evaluation reduces to a standard neural-network forward pass.

Consequently, the inference cost is extremely low. The trained surrogate requires only \(0.31\) ms to generate a complete 3-D temperature-field prediction, corresponding to approximately \(0.52~\mu{\rm s}\) per spatial query point. This computational efficiency enables real-time parametric exploration and interactive material optimization while retaining high-fidelity thermal-field reconstruction. The proposed 4-D VMLFN surrogate therefore provides a practical route toward rapid material design, uncertainty analysis, and thermal-management optimization over continuous physical-parameter spaces.

\subsection{Case V: 3-D Thermoelastic Warping in Heterogeneous Composites}
\label{subsec:case_thermoelastic_warping}

\begin{table}
\centering
\caption{Quantitative Performance Comparison on 3D Thermoelastic Warping (Heterogeneous Composite). Evaluated against FEM benchmark.}
\label{tab:warping_results}
\resizebox{\linewidth}{!}{\begin{tabular}{ccccccc}
\toprule
\textbf{\makecell{Source\\Type}} & \textbf{Method} & \textbf{\makecell{MAX Diff\\(1e-5mm)}} &  \textbf{\makecell{Avg Diff\\ (1e-5mm)}} & \textbf{L2(\%)}& \textbf{Time (s)} & \textbf{Speedup} \\ \midrule
\multirow{3}{*}{Case1} & Comsol & $-$ & $-$ & $-$ &$19$ & $1\times$ \\
& Var. PINN & $5.38$  & $0.96$ & $1.19  $ & $876.5$ & $0.0216\times$ \\
& VMLFN & $\mathbf{15.2}$ &  $\mathbf{1.74}$ & $\mathbf{2.87}$ & $1.26$ &$\mathbf{15.079\times}$ \\ \midrule
\multirow{3}{*}{Case2} & Comsol & $-$ & $-$ & $-$ &$19$ & $1\times$ \\
& Var. PINN & $4.67$  & $1.08$ & $2.9$ & $894.9$& $0.0212\times$ \\
& VMLFN & $\mathbf{34}$ &  $\mathbf{6.97}$ & $\mathbf{2.94}$& $\mathbf{1.36}$ & $\mathbf{13.970\times}$ \\ \midrule
\multirow{3}{*}{Case3} & Comsol & $-$ & $-$ & $-$ &$19$ & $1\times$ \\
& Var. PINN & $6.39$  & $1.67$ & $0.88$ &$899.6$ &  $0.0211\times$ \\
& VMLFN & $\mathbf{16.87}$ &  $\mathbf{7.12}$ & $\mathbf{3.28}$& $\mathbf{1.35}$ & $\mathbf{14.0740\times}$ \\ \midrule
\multirow{3}{*}{Case4} & Comsol & $-$ & $-$ & $-$ &$22$ & $1\times$ \\
& Var. PINN & $26.45$  & $1.88$ & $0.81$ & $901.7$ &$0.0243\times$ \\
& VMLFN & $\mathbf{24.28}$ &  $\mathbf{7.29}$ & $\mathbf{2.29}$& $\mathbf{1.38}$  & $\mathbf{15.942\times}$ \\ 
\bottomrule
\multicolumn{6}{l}{\footnotesize } \\
\end{tabular}}
\end{table}

Finally, we evaluate the proposed framework on a strongly coupled 3-D multiphysics vector-field problem: thermoelastic warping in a heterogeneous composite structure. The computational domain consists of three distinct structural layers with discontinuous material properties. The Young's modulus and Poisson's ratio are given by $E\in\{400,\;162.7,\;129.6\}~{\rm GPa}$, $\nu\in\{0.14,\;0.28,\;0.28\}$,
respectively. The structure is allowed to expand freely under a complex 3-D temperature-gradient field \(\Delta T(\mathbf{x})\), which induces spatially varying thermoelastic deformation and out-of-plane warpage.

As discussed in Section~2.2.5, the proposed variational formulation naturally enforces traction continuity across material interfaces through the weak form. Therefore, no explicit interface penalty terms are required when treating discontinuous material coefficients. This property is particularly important for heterogeneous composite structures, where accurately resolving interfacial mechanical equilibrium is essential. By contrast, a conventional strong-form PINN would need to simultaneously impose the governing equilibrium equations, displacement or traction boundary conditions, and interfacial continuity constraints through multiple penalty terms. Such a formulation is typically sensitive to loss-weight selection, interface sampling density, and optimization stability.

For this reason, we adopt a variational PINN (Var. PINN), rather than a conventional strong-form PINN, as the primary neural baseline for this case. The Var. PINN provides a more appropriate comparison because it also exploits a weak-form formulation and is better suited to heterogeneous thermoelastic problems.


The predicted 3-D displacement field
$\mathbf{u}=[u_x,u_y,u_z]^T$
is benchmarked against high-fidelity FEM solutions computed by COMSOL. Particular attention is given to the out-of-plane displacement component \(u_z\), since it directly characterizes the warpage response of the composite.

As reported in Table~\ref{tab:warping_results}, the Var. PINN generally achieves lower displacement errors in most test cases. However, this improvement in accuracy is obtained at the cost of substantially longer training time. Specifically, the Var. PINN requires \(876.5\)--\(901.7\) s to complete the computation, whereas the proposed VMLFN requires only \(1.26\)--\(1.38\) s. This corresponds to an acceleration of approximately \(650\)--\(700\times\) over the Var. PINN. Compared with COMSOL, VMLFN achieves a speedup of \(13.97\times\)--\(15.94\times\), while still maintaining reasonable agreement with the FEM reference solution.

It is also worth noting that in Case 4, VMLFN attains a slightly lower maximum displacement error than the Var. PINN, with maximum errors of \(24.28\) and \(26.45\), respectively, in units of \(10^{-5}\) mm. This result indicates that, although Var. PINN is generally more accurate in the average-error sense, VMLFN can provide competitive or even superior worst-case behavior in certain heterogeneous configurations.

Overall, these results demonstrate that VMLFN provides an efficient alternative for 3-D thermoelastic warping analysis in heterogeneous composites. By combining a variational formulation with direct matrix-based learning, the method achieves a favorable trade-off between predictive accuracy and computational cost. It avoids expensive iterative neural-network training while preserving the key physical properties required for multiphysics composite analysis, including natural satisfaction of interfacial traction continuity.

To further illustrate the reconstruction quality, Fig.~\ref{fig:warping_results_all} visualizes the predicted thermoelastic warpage field and the corresponding absolute error distribution across the composite interfaces. The comparison confirms that VMLFN captures the main deformation pattern and interfacial warpage characteristics with high computational efficiency.

\section{Conclusion}
\label{sec:concl}

In this paper, we proposed a variational matrix-learning Fourier network (VMLFN) framework for efficient PDE-governed multiphysics surrogate modeling. The proposed method combines a log-space sine neural-network representation with a variational weak-form formulation, converting the training of output-layer weights into a direct matrix-solving problem through the zero-gradient stationarity condition. Since the hidden-layer Fourier features are fixed and only first-order derivatives are required, VMLFN avoids high-order automatic differentiation, iterative backpropagation, and penalty-coefficient tuning between PDE residuals and boundary conditions. A heuristic frequency-scanning algorithm was further introduced to select a problem-adaptive maximum frequency, enabling the spectral basis to capture the dominant physical modes of the target problem. Dirichlet boundary conditions are embedded into the trial solution, while Neumann boundaries and interface traction continuity are naturally handled by the variational formulation. These properties allow the proposed framework to effectively address complex boundary and heterogeneous-interface conditions without additional penalty-loss balancing. The proposed framework was validated on three representative classes of multiphysics problems, including heat conduction, solid mechanics, and Helmholtz wave propagation, across five benchmark cases. Numerical results show that VMLFN achieves accurate field reconstruction with substantial computational speedup compared with conventional PINNs and repeated FEM simulations. In particular, the proposed 4-D spatio-parametric surrogate enables fast zero-shot prediction for unseen material parameters, supporting real-time parametric design exploration.

\bibliographystyle{ieeetr}
\bibliography{./IEEErefer.bib}

@ARTICLE{Lau:TCPMT22,
  author={Lau, John H.},
  journal={IEEE Transactions on Components, Packaging and Manufacturing Technology}, 
  title={Recent Advances and Trends in Advanced Packaging}, 
  year={2022},
  volume={12},
  number={2},
  pages={228-252}}

@ARTICLE{Giacomini:JESSCDC24,
  author={Giacomini Rocha, Leandro M. and Naeim, Mohamed and Paim, Guilherme and Brunion, Moritz and Venugopal, Priya and Milojevic, Dragomir and Myers, James and Badaroglu, Mustafa and Verhelst, Marian and Ryckaert, Julien and Biswas, Dwaipayan},
  journal={IEEE Journal on Exploratory Solid-State Computational Devices and Circuits}, 
  title={System-Technology Co-Optimization for Dense Edge Architectures Using 3-D Integration and Nonvolatile Memory}, 
  year={2024},
  volume={10},
  number={},
  pages={125-134}}

@INPROCEEDINGS{Biswas:ISVLSI24,
  author={Biswas, Dwaipayan and Myers, James and Samavedam, Srikanth B. and Ryckaert, Julien},
  booktitle={2024 IEEE Computer Society Annual Symposium on VLSI (ISVLSI)}, 
  title={STCO: Driving the More than Moore Era}, 
  year={2024},
  volume={},
  number={},
  pages={7-8}}

@INPROCEEDINGS{Wang:IEDM24,
  author={Wang, Linqiu and Xie, Feifan and Liu, Jizhe and Liu, Tianchi and Peng, Lianmao and Zhang, Zhiyong and Wei, Tiwei and Chen, Rongmei},
  booktitle={2024 IEEE International Electron Devices Meeting (IEDM)}, 
  title={Power and Thermal Integrity Analysis of High Performance and Low Power CPUs at Sub-2nm Node Designed with Various Advanced Backside PDNs}, 
  year={2024},
  volume={},
  number={},
  pages={1-4}}

@INPROCEEDINGS{Parekh:ICCD25,
  author={Parekh, Varun Darshana and Hazenstab, Zachary Wyatt and Srinivasa, Srivatsa Rangachar and Chakrabarty, Krishnendu and Ni, Kai and Narayanan, Vijaykrishnan},
  booktitle={2025 IEEE 43rd International Conference on Computer Design (ICCD)}, 
  title={STAMP-2.5D: Structural and Thermal Aware Methodology for Placement in 2.5D Integration}, 
  year={2025},
  volume={},
  number={},
  pages={159-166}}

@ARTICLE{Abouelyazid:tcad22,
  author={Abouelyazid, Mohamed Saleh and Hammouda, Sherif and Ismail, Yehea},
  journal={IEEE Transactions on Computer-Aided Design of Integrated Circuits and Systems}, 
  title={Accuracy-Based Hybrid Parasitic Capacitance Extraction Using Rule-Based, Neural-Networks, and Field-Solver Methods}, 
  year={2022},
  volume={41},
  number={12},
  pages={5681-5694}}

@INPROCEEDINGS{Zhu:DATE25,
  author={Zhu, Tianxiang and Wang, Qipan and Lin, Yibo and Wang, Runsheng and Huang, Ru},
  booktitle={2025 Design, Automation \& Test in Europe Conference (DATE)}, 
  title={MORE-Stress: Model Order Reduction based Efficient Numerical Algorithm for Thermal Stress Simulation of TSV Arrays in 2.5D/3D IC}, 
  year={2025},
  volume={},
  number={},
  pages={1-7}}

@ARTICLE{Li:TGRS26,
  author={Li, Anyu},
  journal={IEEE Transactions on Geoscience and Remote Sensing}, 
  title={Efficient One-Way Wave-Equation Depth Migration Using Fast Fourier Transform and Complex Padé Approximation via Helmholtz Operator}, 
  year={2026},
  volume={64},
  number={},
  pages={1-11}}

@InProceedings{Long:ICML18,
  title = 	 {{PDE}-Net: Learning {PDE}s from Data},
  author =       {Long, Zichao and Lu, Yiping and Ma, Xianzhong and Dong, Bin},
  booktitle = 	 {Proceedings of the 35th International Conference on Machine Learning},
  pages = 	 {3208--3216},
  year = 	 {2018},
  editor = 	 {Dy, Jennifer and Krause, Andreas},
  volume = 	 {80},
  series = 	 {Proceedings of Machine Learning Research},
  month = 	 {10--15 Jul},
  publisher =    {PMLR}
}

@article{Lu:nmi21,
  title   = {Learning nonlinear operators via {DeepONet} based on the universal approximation theorem of operators},
  author  = {Lu, Lu and Jin, Pengzhan and Pang, Guofei and Zhang, Zhongqiang and Karniadakis, George Em},
  journal = {Nature Machine Intelligence},
  volume  = {3},
  number  = {3},
  pages   = {218--229},
  year    = {2021}
}

@inproceedings{
li:ICLR21,
title={Fourier Neural Operator for Parametric Partial Differential Equations},
author={Zongyi Li and Nikola Borislavov Kovachki and Kamyar Azizzadenesheli and Burigede liu and Kaushik Bhattacharya and Andrew Stuart and Anima Anandkumar},
booktitle={International Conference on Learning Representations},
year={2021}
}

@ARTICLE{Cao:NMI24,
  author={Qianying Cao and Somdatta Goswami and George Em Karniadakis},
  journal={IEEE Trans. Neural Netw. Learn. Syst.}, 
  title={Laplace neural operator for solving differential equations}, 
  year={2024},
  number={6},
  volume={6},
  pages={631–640}}

@ARTICLE{Zhang:TPAMI25,
  author={Zhang, Rui and Meng, Qi and Zhu, Rongchan and Wang, Yue and Shi, Wenlei and Zhang, Shihua and Ma, Zhi-Ming and Liu, Tie-Yan},
  journal={IEEE Transactions on Pattern Analysis and Machine Intelligence}, 
  title={Monte Carlo Neural PDE Solver for Learning PDEs via Probabilistic Representation}, 
  year={2025},
  volume={47},
  number={6}}

@ARTICLE{Huang:TNNLS25,
  author={Huang, Shudong and Feng, Wentao and Tang, Chenwei and He, Zhenan and Yu, Caiyang and Lv, Jiancheng},
  journal={IEEE Trans. Neural Netw. Learn. Syst.}, 
  title={Partial Differential Equations Meet Deep Neural Networks: A Survey}, 
  year={2025},
  volume={36},
  number={8},
  pages={13649-13669}}

@article{20DEM,
title = {An energy approach to the solution of partial differential equations in computational mechanics via machine learning: Concepts, implementation and applications},
journal = {Comput. Methods Appl. Mech. Engrg.},
volume = {362},
pages = {112790},
year = {2020},
issn = {0045-7825},
doi = {https://doi.org/10.1016/j.cma.2019.112790},
author = {E. Samaniego and C. Anitescu and S. Goswami and V.M. Nguyen-Thanh and H. Guo and K. Hamdia and X. Zhuang and T. Rabczuk},
}

@article{2020xpinn,
  title={Extended Physics-informed Neural Networks (XPINNs): A Generalized Space-Time Domain Decomposition based Deep Learning Framework for Nonlinear Partial Differential Equations},
  journal={Commun. Comput. Phys.},
  volume = {28(5)},
  year={2020},
  issn = {2002-2041},
  doi={https://doi.org/10.4208/cicp.OA-2020-0164},
  author={Ameya Dilip Jagtap and George E. Karniadakis},
}

@article{20cpinn,
title = {Conservative physics-informed neural networks on discrete domains for conservation laws: Applications to forward and inverse problems},
journal = {Comput. Methods Appl. Mech. Engrg.},
volume = {365},
pages = {113028},
year = {2020},
issn = {0045-7825},
doi = {https://doi.org/10.1016/j.cma.2020.113028},
author = {Ameya D. Jagtap and Ehsan Kharazmi and George E. Karniadakis},
}

@ARTICLE{Wang:TNNLS25,
  author={Wang, Honghui and Pu, Yifan and Song, Shiji and Huang, Gao},
  journal={IEEE Transactions on Neural Networks and Learning Systems}, 
  title={Advancing Generalization in PINNs Through Latent-Space Representations}, 
  year={2025},
  volume={36},
  number={12},
  pages={20243-20257}}

@ARTICLE{Luca:TNNLS25,
  author={Menicali, Luca and Richter, David H. and Castruccio, Stefano},
  journal={IEEE Transactions on Neural Networks and Learning Systems}, 
  title={Bayesian Neural Networks With Physics-Informed Priors With Application to Boundary Layer Velocity}, 
  year={2025},
  volume={36},
  number={10},
  pages={18048-18061}}

@ARTICLE{Wong:TAI24,
  author={Wong, Jian Cheng and Ooi, Chin Chun and Gupta, Abhishek and Ong, Yew-Soon},
  journal={IEEE Transactions on Artificial Intelligence}, 
  title={Learning in Sinusoidal Spaces With Physics-Informed Neural Networks}, 
  year={2024},
  volume={5},
  number={3},
  pages={985-1000}}

@ARTICLE{Xiao:TII26,
  author={Xiao, Guolin and Lang, Qi and Lu, Wei and Liu, Xiaodong},
  journal={IEEE Transactions on Industrial Informatics}, 
  title={Physics-Guided Spectral Neural Networks With Self-Discovered Partial Differential Equations for Sparse Field Reconstruction}, 
  year={2026},
  volume={22},
  number={3},
  pages={1771-1781}}

@ARTICLE{Song:TGRS24,
  author={Song, Chao and Zhao, Tianshuo and Bin Waheed, Umair and Liu, Cai and Tian, You},
  journal={IEEE Transactions on Geoscience and Remote Sensing}, 
  title={Seismic Traveltime Simulation for Variable Velocity Models Using Physics-Informed Fourier Neural Operator}, 
  year={2024},
  volume={62},
  number={},
  pages={1-9}}

@ARTICLE{Qi:TMTT25,
  author={Qi, Shutong and Sarris, Costas D.},
  journal={IEEE Transactions on Microwave Theory and Techniques}, 
  title={Physics-Informed Deep Operator Network for 3-D Time-Domain Electromagnetic Modeling}, 
  year={2025},
  volume={73},
  number={7},
  pages={3800-3812}}

@article{Panghal:EC21,
title = {Optimization free neural network approach for solving ordinary and partial differential equations},
journal = {Engineering with Computers},
volume = {37},
number={4},
pages = {2989-3002},
year = {2021},
author = {Panghal, Shagun and Kumar, Manoj},
}

@article{Sun:NPL19,
title = {Solving Partial Differential Equation Based on Bernstein Neural Network and Extreme Learning Machine Algorithm},
journal = {Neural Processing Letters},
volume = {50},
number={2},
pages = {1153-1172},
year = {2019},
author = {Sun, Hongli and Hou, Muzhou and Yang, Yunlei and Zhang, Tianle and Weng, Futian and Han, Feng}
}

@article{19PINN,
title = {Physics-informed neural networks: A deep learning framework for solving forward and inverse problems involving nonlinear partial differential equations},
journal = {J. Comput. Phys.},
volume = {378},
pages = {686-707},
year = {2019},
issn = {0021-9991},
doi = {https://doi.org/10.1016/j.jcp.2018.10.045},
author = {M. Raissi and P. Perdikaris and G.E. Karniadakis},
}

@article{20PIELM,
title = {Physics Informed Extreme Learning Machine (PIELM)–A rapid method for the numerical solution of partial differential equations},
journal = {Neurocomputing},
volume = {391},
pages = {96-118},
year = {2020},
issn = {0925-2312},
doi = {https://doi.org/10.1016/j.neucom.2019.12.099},
author = {Vikas Dwivedi and Balaji Srinivasan},
}

@article{23binn,
title = {BINN: A deep learning approach for computational mechanics problems based on boundary integral equations},
journal = {Comput. Methods Appl. Mech. Engrg.},
volume = {410},
pages = {116012},
year = {2023},
issn = {0045-7825},
doi = {https://doi.org/10.1016/j.cma.2023.116012},
author = {Jia Sun and Yinghua Liu and Yizheng Wang and Zhenhan Yao and Xiaoping Zheng},
}

@article{21lelmm,
title = {Local extreme learning machines and domain decomposition for solving linear and nonlinear partial differential equations},
journal = {Comput. Methods Appl. Mech. Engrg.},
volume = {387},
pages = {114129},
year = {2021},
issn = {0045-7825},
doi = {https://doi.org/10.1016/j.cma.2021.114129},
author = {Suchuan Dong and Zongwei Li},
}

@article{2021elmforpde,
  author = {Fabiani, Gianluca and Calabr{\`o}, Francesco and Russo, Lucia and Siettos, Constantinos},
  title = {Numerical solution and bifurcation analysis of nonlinear partial differential equations with extreme learning machines},
  journal = {Journal of Scientific Computing},
  year = {2021},
  volume = {89},
  number = {2},
  pages = {44},
  issn = {1573-7691},
  doi = {10.1007/s10915-021-01650-5},
}

@article{23bayesianELM,
title = {Bayesian physics-informed extreme learning machine for forward and inverse PDE problems with noisy data},
journal = {Neurocomputing},
volume = {549},
pages = {126425},
year = {2023},
issn = {0925-2312},
doi = {https://doi.org/10.1016/j.neucom.2023.126425},
author = {Xu Liu and Wen Yao and Wei Peng and Weien Zhou},
}

@article{24hdimELM,
title = {An extreme learning machine-based method for computational PDEs in higher dimensions},
journal = {Comput. Methods Appl. Mech. Engrg.},
volume = {418},
pages = {116578},
year = {2024},
issn = {0045-7825},
doi = {https://doi.org/10.1016/j.cma.2023.116578},
author = {Yiran Wang and Suchuan Dong},
}

@article{Chen:JML22,
author = {Jingrun Chen and Xurong Chi and Weinan E and Zhouwang Yang},
title={Bridging Traditional and Machine Learning-Based Algorithms for Solving PDEs: The Random Feature Method}, 
volume={1}, 
number={3}, 
journal={Journal of Machine Learning}, 
year={2022}, 
month={Sept.}, 
pages={268–298}}

@article{Davide:jcp26,
title = {Least squares with equality constraints extreme learning machines for the resolution of PDEs},
journal = {Journal of Computational Physics},
volume = {547},
pages = {114553},
year = {2026},
issn = {0021-9991},
author = {Davide {Elia De Falco} and Enrico Schiassi and Francesco Calabrò}
}

@article{Yang:SC20,
author = {Yang, Yunlei and Hou, Muzhou and Sun, Hongli and Zhang, Tianle and Weng, Futian and Luo, Jianshu},
title = {Neural network algorithm based on Legendre improved extreme learning machine for solving elliptic partial differential equations},
year = {2020},
issue_date = {Jan 2020},
publisher = {Springer-Verlag},
address = {Berlin, Heidelberg},
volume = {24},
number = {2},
issn = {1432-7643},
journal = {Soft Comput.},
month = jan,
pages = {1083–1096},
numpages = {14}
}

@article{Shivani:NS26,
title = {Numerical solution of elliptic PDEs via Vieta–Fibonacci wavelet based neural networks},
journal = {Nonlinear Science},
volume = {7},
pages = {100137},
year = {2026},
issn = {3050-5178},
author = {Shivani Aeri and Rakesh Kumar}
}

@article{Wang:IJCM26,
author = {Wansheng Wang and Jiangtao Pan and Feifei Gao and Hongxuan Liu and Lijia Zhou},
title = {Laguerre neural network for solving neutral delay differential equations based on the extreme learning machine},
journal = {International Journal of Computer Mathematics},
volume = {103},
number = {2},
pages = {357--378},
year = {2026},
publisher = {Taylor \& Francis},
}

@article{Enrico:N21,
title = {Extreme theory of functional connections: A fast physics-informed neural network method for solving ordinary and partial differential equations},
journal = {Neurocomputing},
volume = {457},
pages = {334-356},
year = {2021},
issn = {0925-2312},
author = {Enrico Schiassi and Roberto Furfaro and Carl Leake and Mario {De Florio} and Hunter Johnston and Daniele Mortari}
}

@INPROCEEDINGS{Chen:ICCAD23,
  author={Chen, Liang and Lu, Jincong and Jin, Wentian and Tan, Sheldon X.-D.},
  booktitle={2023 IEEE/ACM International Conference on Computer Aided Design (ICCAD)}, 
  title={Fast Full-Chip Parametric Thermal Analysis Based on Enhanced Physics Enforced Neural Networks}, 
  year={2023},
  volume={},
  number={},
  pages={1-8}}

@ARTICLE{Chen:TCAD25,
  author={Chen, Liang and Zhu, Wenxing and Tang, Min and Tan, Sheldon X.-D. and Mao, Jun-Fa and Zhang, Jianhua},
  journal={IEEE Transactions on Computer-Aided Design of Integrated Circuits and Systems}, 
  title={PISOV: Physics-Informed Separation of Variables Solvers for Full-Chip Thermal Analysis}, 
  year={2025},
  volume={44},
  number={5},
  pages={1874-1886}}

@ARTICLE{Chen:TCAD23,
  author={Chen, Liang and Jin, Wentian and Kavousi, Mohammadamir and Lamichhane, Subed and Tan, Sheldon X.-D.},
  journal={IEEE Transactions on Computer-Aided Design of Integrated Circuits and Systems}, 
  title={Linear Time Electromigration Analysis Based on Physics-Informed Sparse Regression}, 
  year={2023},
  volume={42},
  number={11},
  pages={4126-4138}}

@article{Lu:SIAM21,
author = {Lu, Lu and Pestourie, Rapha\"{e}l and Yao, Wenjie and Wang, Zhicheng and Verdugo, Francesc and Johnson, Steven G.},
title = {Physics-Informed Neural Networks with Hard Constraints for Inverse Design},
journal = {SIAM Journal on Scientific Computing},
volume = {43},
number = {6},
pages = {B1105-B1132},
year = {2021}
}

@misc{Ren:25,
      title={General Fourier Feature Physics-Informed Extreme Learning Machine (GFF-PIELM) for High-Frequency PDEs}, 
      author={Fei Ren and Sifan Wang and Pei-Zhi Zhuang and Hai-Sui Yu and He Yang},
      year={2025},
      eprint={2510.12293},
      archivePrefix={arXiv},
      primaryClass={cs.LG}
}

@article{Ngom:SADM21,
author = {Ngom, Marieme and Marin, Oana},
title = {Fourier neural networks as function approximators and differential equation solvers},
journal = {Statistical Analysis and Data Mining: The ASA Data Science Journal},
volume = {14},
number = {6},
pages = {647-661},
year = {2021}
}

\end{document}